\documentclass[sigconf]{acmart}

\copyrightyear{2025}
\acmYear{2025}
\setcopyright{cc}
\setcctype{by}
\acmConference[MM '25]{Proceedings of the 33rd ACM International Conference on Multimedia}{October 27--31, 2025}{Dublin, Ireland}
\acmBooktitle{Proceedings of the 33rd ACM International Conference on Multimedia (MM '25), October 27--31, 2025, Dublin, Ireland}\acmDOI{10.1145/3746027.3758180}
\acmISBN{979-8-4007-2035-2/2025/10}

\usepackage{graphicx}       
\usepackage{array}           
\usepackage{booktabs} 
\usepackage{colortbl} 
\usepackage{multirow}        
\usepackage{hhline}          
\usepackage{caption}        
\usepackage{tabularx} 
\usepackage{booktabs}
\usepackage{makecell}
\usepackage{enumitem}
\usepackage{tcolorbox}
\usepackage{booktabs}
\usepackage{adjustbox}
\usepackage{fancyhdr}
\usepackage{fontawesome}

\makeatletter
\begin{document}
	
	\title[AirScape: An Aerial Generative World Model with Motion Controllability]{AirScape: An Aerial Generative World Model with\\ Motion Controllability}
	

	\author{Baining Zhao}
	\authornote{All four authors contributed equally to this research. Rongze Tang and Ziyou Wang conducted this work during their internship at Tsinghua University.}
	\affiliation{%
		\institution{Shenzhen International Graduate School, Tsinghua University \\ Pengcheng Laboratory}
		\city{Shenzhen}
		\country{China}}
	
	\author{Rongze Tang}
	\authornotemark[1]
	\affiliation{%
		\institution{School of Computer Science \& Technology, Beijing Institute of Technology}
		\city{Beijing}
		\country{China}}
		
        \author{Mingyuan Jia}
	\authornotemark[1]
	\affiliation{%
		\institution{Department of Automation, Tsinghua University}
		\city{Beijing}
		\country{China}}
	
	\author{Ziyou Wang}
        \authornotemark[1]
	\affiliation{%
		\institution{Computer and Communication Engineering College, Northeastern University}
		\city{Qinhuangdao}
		\country{China}}
	
	\author{Fanhang Man}
	\affiliation{%
	\institution{Shenzhen International Graduate School, Tsinghua University}
	\city{Shenzhen}
	\country{China}}
	
	\author{Xin Zhang}
	\affiliation{%
		\institution{Manifold AI}
		\city{Beijing}
		\country{China}}
		
	\author{Yu Shang}
	\affiliation{%
		\institution{Department of Electronic Engineering, Tsinghua University}
		\city{Beijing}
		\country{China}}
		
	\author{Weichen Zhang}
	\affiliation{%
		\institution{Shenzhen International Graduate School, Tsinghua University \\ Pengcheng Laboratory}
		\city{Shenzhen}
		\country{China}}
		
	\author{Wei Wu}
	\affiliation{%
		\institution{Manifold AI}
		\city{Beijing}
		\country{China}}
	
	\author{Chen Gao}
	\authornote{Corresponding authors: Chen Gao (chgao96@gmail.com), Xinlei Chen (chen.xinlei@sz.tsinghua.edu.cn), Yong Li (liyong07@tsinghua.edu.cn)}
	\affiliation{%
		\institution{BNRist, Tsinghua University}
		\city{Beijing}
		\country{China}}
	
	\author{Xinlei Chen}
	\authornotemark[2]
	\affiliation{%
		\institution{Shenzhen International Graduate School, Tsinghua University}
		\city{Shenzhen}
		\country{China}}
	
	\author{Yong Li}
        \authornotemark[2]
	\affiliation{%
		\institution{Department of Electronic Engineering, Tsinghua University \\ BNRist, Tsinghua University}
		\city{Beijing}
		\country{China}}

	\renewcommand{\shortauthors}{Baining Zhao et al.}
	\renewcommand{\shorttitle}{AirScape: An Aerial Generative World Model with Motion Controllability}

	\begin{abstract}

How to enable agents to predict the outcomes of their own motion intentions in three-dimensional space has been a fundamental problem in embodied intelligence. To explore general spatial imagination capability, we present AirScape, the first world model designed for six-degree-of-freedom aerial agents. AirScape predicts future observation sequences based on current visual inputs and motion intentions. Specifically, we construct a dataset for aerial world model training and testing, which consists of 11k video-intention pairs. This dataset includes first-person-view videos capturing diverse drone actions across a wide range of scenarios, with over 1,000 hours spent annotating the corresponding motion intentions. Then we develop a two-phase schedule to train a foundation model—initially devoid of embodied spatial knowledge—into a world model that is controllable by motion intentions and adheres to physical spatio-temporal constraints. Experimental results demonstrate that AirScape significantly outperforms existing foundation models in 3D spatial imagination capabilities, especially with over a 50\% improvement in metrics reflecting motion alignment. The project is available at: \textcolor{blue}{\url{https://embodiedcity.github.io/AirScape/}}.
\end{abstract}


	\maketitle

\section{Introduction}
\label{sec:intro}

\begin{figure*}[h]
	\centering
    \vspace{-20pt}
	\includegraphics[width= \textwidth]{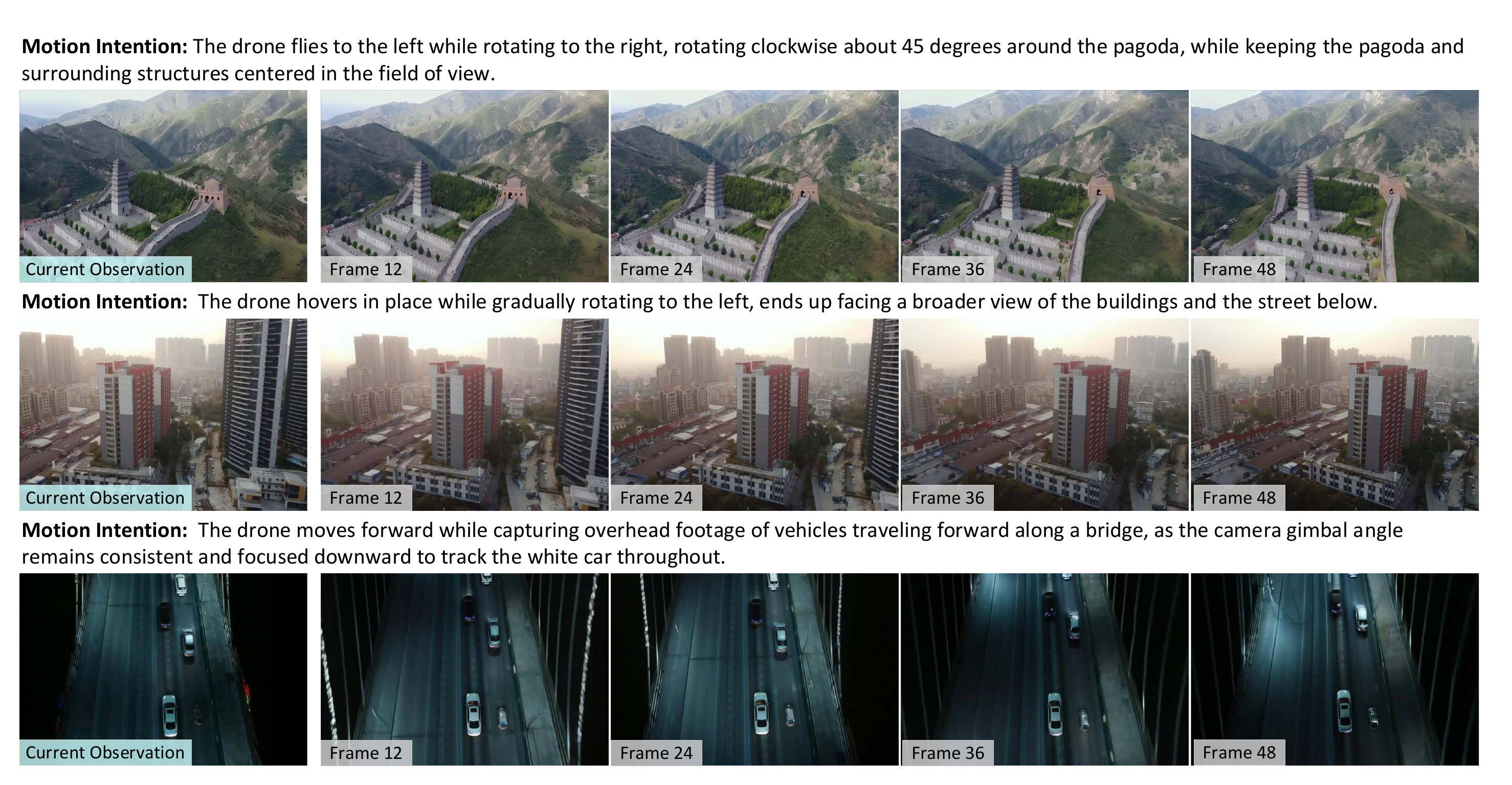}
	\vspace{-23pt}
	\caption{In 3D space, AirScape can predict the sequence of observations that would result if a six-degree-of-freedom aerial agent executed a series of actions to achieve an intention, based on current visual observations. AirScape can handle diverse actions (translation, rotation, and their combinations), environments (rural, urban), viewpoints (top-down, horizon), and lighting conditions (daytime, dusk, nighttime), simulating embodied observation characteristics such as perspective and parallax.}
	\label{fig:task}
    \vspace{-10pt}
\end{figure*}

The unprecedented advancements in generative models~\cite{cao2024survey} have catalyzed a paradigm shift in the development of world models, enabling generation and simulation of the real-world environment, with inputs of texts and actions.
The simulation reflects a kind of high-level capability, counterfactual reasoning, which enables the simulation and prediction of possible outcomes based on hypothetical conditions or decisions~\cite{ding2024understanding}. By comparing results under different assumptions, world models support better decision-making in unknown or complex environments, which is particularly important for downstream applications such as embodied robotics~\cite{wu2023daydreamer,li2025robotic}, autonomous driving~\cite{gao2023magicdrive,gao2024magicdrive3d}, etc. 
Moreover, the world model can enhance the agents' spatial intelligence~\cite{gupta2021embodied,zha2025enable}, with the human-like critical ability in understanding the environment.
Specifically, when an agent operates in a three-dimensional real-world space, given its current observation, we expect it to predict how its first-person perspective will change after performing actions or tasks, which implicitly indicates how its spatial relationship with the surrounding environment will evolve. This goes beyond basic spatial perception and understanding~\cite{zhao2025embodiedr,cheng2024videollama}, enabling navigation~\cite{an2024etpnav,zhang2025citynavagent} and task planning~\cite{panowicz2024robust} in complex and unknown real-world scenarios.

Current research on spatial world models primarily focuses on humanoid robots and autonomous driving applications~\cite{guan2024world,sakagami2023robotic}. However, the world models for humanoid robots emphasize manipulation and indoor environment modeling, while those for autonomous driving focus on predicting driving behaviors and modeling road dynamics. Both of them operate mostly in two-dimensional planes with limited action spaces. 
With the development of low-altitude economy, the increasing intelligence of aerial agents, such as drones, drive their widespread applications, such as delivery~\cite{chen2024ddl,liu2025meal}, emergency disaster relief~\cite{chen2024soscheduler,schedl2021autonomous,xu2024scalable}, and urban pollution management~\cite{zhou2025catua,rafael2020autonomous, liu2024mobiair}.
The research on aerial world models remains unexplored. Moreover, the spatial geometric complexity of embodied spatial counterfactual reasoning in 3D real-world environments with six degrees of freedom (6DoF) is significantly higher, representing a more general type of world model. Examples of the aerial world model are presented in Figure \ref{fig:task}.

Vision is one of the fundamental perceptual modalities, and the latest visual observations inherently contain spatial information~\cite{zhao2025urbanvideobench}. Compared to flight control variables or specific trajectory coordinates, we argue that expressing action intentions in textual form aligns more closely with human reasoning processes and offers greater flexibility. Text-based instructions can represent high-level navigation instructions, such as "move to the boat ahead" or "follow the car in front," as well as low-level commands, such as "rotate 90 degrees to the left." When a series of actions is executed in the current spatial context to fulfill an intention, the most direct outcome is a sequence of visual observations. This input-output structure aligns with the framework of video generation models conditioned on both graphical and textual inputs.
Recently, video generation foundation models, represented by diffusion~\cite{NEURIPS2020_4c5bcfec,blattmann2023stable} and autoregressive models~\cite{chen2020generative,yu2022scaling}, have rapidly advanced and are becoming important tools for implementing world models~\cite{lecun2022path,agarwal2025cosmos}. Inspired by the scaling law, large foundation models exhibit generalization capabilities~\cite{peebles2023scalable,kong2024hunyuanvideo}. Video generation models can model dynamic changes in temporal sequences, directly simulating visual information of the environment. This capability aligns closely with the requirements of world models for modeling and predicting future spatio-temporal states. However, constructing a generative aerial world model still faces the following challenges:

\begin{itemize}[leftmargin=*]
\vspace{-5pt}
	\item \textbf{Lack of aerial datasets}: Training world models requires first-person perspective videos and corresponding textual prompts about aerial agents' actions or tasks. Existing datasets are either third-person views or ground-based perspectives from robots or vehicles~\cite{hu2022make,caesar2020nuscenes,dasari2019robonet}.
	\item \textbf{Distribution gap between video foundation models and world models}: In terms of text input, existing open-source foundation models focus on generating videos from detailed textual descriptions~\cite{ho2022video,blattmann2023stable}, whereas world models rely on concise instructions or action intents. In terms of video, training data for open-source foundation models mostly consists of third-person videos with limited visual changes~\cite{yang2024cogvideox,kong2024hunyuanvideo,kondratyuk2023videopoet}, while embodied first-person perspectives typically have narrower fields of view and larger visual changes, increasing training difficulty.
	\item \textbf{Diversity in generation}: Drones operate in 6DoF with high flexibility~\cite{wang2025ultra}. Compared to ground vehicles, generated scenes include lateral translation, in-place rotation, camera gimbal adjustments, and combinations of multiple actions, making generation more challenging. The aerial spatial world model is required to simulate more complex changes in relative position, perspective variation, and parallax effects.
\end{itemize}

To address these issues, we first introduce an 11k dataset for training aerial world models. We collect videos from three public drone datasets, segment and filter them, and annotate each video clip with its corresponding motion intents using large multimodal models (LMMs) and human refinement (\textbf{Section \ref{sec:dataset}}). Subsequently, we develop a two-stage training schedule: fine-tuning video generation foundation models to adapt to the text and video distributions; rejection sampling and self-play training are employed to further improve generation outputs that violate spatial physical constraints. (\textbf{Section \ref{sec:method}}). Experimental results demonstrate that the proposed spatial world model can predict observations in embodied perspectives when performing various actions or tasks (\textbf{Section \ref{sec:experiments}}). The main contributions of this paper are as follows:

\begin{itemize}[leftmargin=*]
	\item \textbf{The first dataset for training and testing generative world models in aerial spaces}, containing \textbf{11k} video clips with corresponding textual motion intentions.
	\item \textbf{The first generative world model in aerial spaces}, capable of predicting visual observations from controllable motion intentions in three-dimensional spaces.
	\item Experimental analysis demonstrates that our proposed AirScape exhibits embodied motion-following simulation and prediction capabilities in aerial space scenarios, outperforming existing general video generation models and world models.
\end{itemize}

	\section{Related Work}

\noindent \textbf{World Model.} 
World models present a grand vision, serving as simulators to support offline training and interaction for agents, while also enabling high-level reasoning and generalization in real-time decision-making~\cite{ha2018world,lecun2022path,agarwal2025cosmos}. Current research on world models can be categorized into several key areas. First, general world models aim to develop scalable and generalizable representations to simulate and understand complex environments~\cite{wang2024worlddreamer,bruce2024genie,tot2025adapting}. Second, world models for embodied AI focus on enabling robots to learn and construct world models through interaction with their surroundings, improving manipulation and navigation capabilities~\cite{gu2024advancing,zhou2024robodreamer}. Third, applications in autonomous driving utilize world models to simulate traffic scenarios and enhance driving safety~\cite{gao2023magicdrive,wang2024drivedreamer,russell2025gaia}. 
However, existing world model research has not yet focused on aerial agents~\cite{gao2024embodiedcity,wang2024transformloc}. Actually, the aerial agents have the potential to exhibit more generalized spatial intelligence due to their six degrees of freedom in three-dimensional space.

\noindent \textbf{Video Generation.}
Exemplified by Sora~\cite{sora2023}, video generation models have garnered significant attention for their highly realistic and lifelike video generation capabilities~\cite{croitoru2023diffusion}. Compared to GAN-based approaches~\cite{chu2020learning}, diffusion-based models have demonstrated superior performance in generating high-fidelity videos~\cite{NEURIPS2020_4c5bcfec,blattmann2023stable,hong2022cogvideo}. Additionally, inspired by the success of transformer architectures in large language models (LLMs), several works have explored autoregressive-based approaches for video generation~\cite{liu2024mardini,chen2020generative,yu2022scaling}, leveraging their sequential modeling capabilities.
In terms of applications, video generation has expanded into diverse directions. Text-to-video generation has made significant strides, with works like Imaginaire setting new benchmarks for producing high-quality videos from textual prompts~\cite{wan2025wan,skorokhodov2024hierarchical}. Image-to-video approaches focus on animating static images based on motion descriptions or physical constraints, enabling dynamic visualizations from static inputs~\cite{zhao2024identifying,jiang2024videobooth}.
Despite these advancements, current video generation models are primarily designed to visualize input content, relying heavily on detailed descriptions to control video outputs. They often lack the predictive and reasoning capabilities inherent to world models, which are essential for understanding and simulating the consequences of actions, particularly in complex environments like aerial spaces with six degrees of freedom.

\section{Dataset for Aerial World Model}
\label{sec:dataset}

We present an 11k embodied aerial agent video dataset along with corresponding annotations of motion intention, aligning the inputs and outputs of the aerial world model. Below, we detail the dataset construction pipeline and dataset statistics.

\begin{figure*}[t]
	\centering
    \vspace{-10pt}
	\includegraphics[width = 0.8\linewidth]{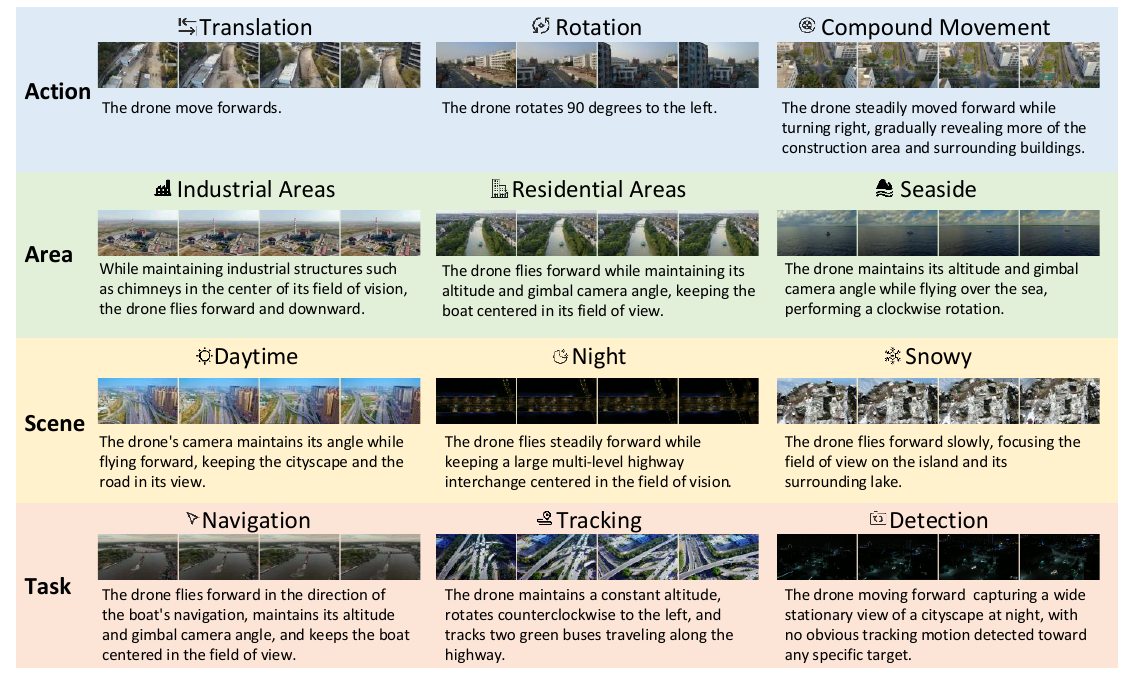}
	\vspace{-10pt}
	\caption{The proposed dataset includes samples with diverse actions, areas, scenes, and tasks.}
	\label{Fig:example}
	\vspace{-10pt}
\end{figure*}

\subsection{Dataset Construction Pipeline}
We first gather the egocentric perspective videos shot by the UAVs from open-sourced dataset: UrbanVideo-Bench~\cite{zhao2025urbanvideobench}, NAT2021~\cite{ye2022unsupervised}, and WebUAV-3M~\cite{10004511}. These datasets are derived from various tasks, including vision-language navigation, tracking, etc., featuring diverse drone actions. The scenes span over 10 types, such as industrial areas, residential zones, suburbs, and coastal regions. Additionally, they include various weather conditions, such as sunny days and nighttime. We segmented the videos into 129-frame clips and filtered out those that were static or exhibited abrupt changes. Examples are presented in Figure \ref{Fig:example}.

Subsequently, we employed a vision-language model to infer the drone's motion intentions. These intentions could range from simple individual actions (e.g., rotating 45 degrees to the left) to specific tasks (e.g., tracking the white car ahead). We designed a straightforward chain-of-thought process, first identifying the action, then summarizing its stopping conditions, and finally merging them into a coherent and logically structured intention prompt.

Finally, we conducted over 1,000 hours of human refinement. Even the most advanced VLMs struggle to accurately infer the agent's motion from changes in the embodied perspective. Therefore, we focused on correcting the following aspects: incorrect actions, ambiguous descriptions, and imprecise tasks within the descriptions. The pipeline is shown in Figure \ref{Fig:statistic}a.

\subsection{Dataset Statistics}

\begin{figure*}[t]
	\centering
	\includegraphics[width = \linewidth]{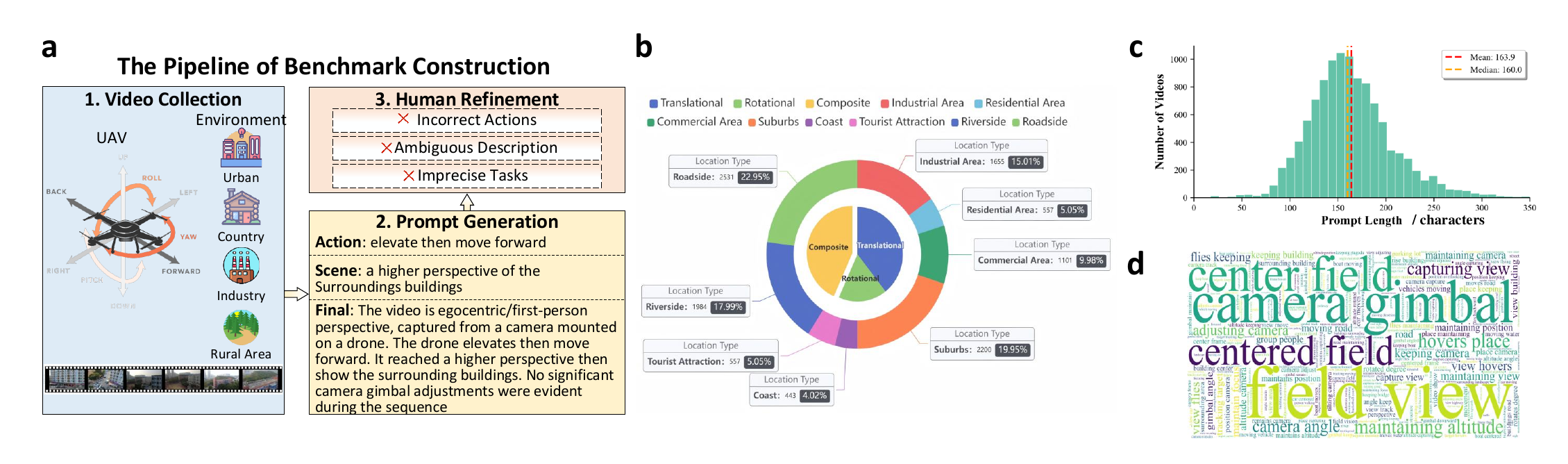}
	\vspace{-30pt}
	\caption{\textbf{a.} Dataset construction pipeline. \textbf{b.} Proportions of different actions and various scenarios in the dataset. \textbf{c.} Length distribution of intention prompts in the dataset. \textbf{d.} Word cloud of intention prompts in the dataset.}
	\label{Fig:statistic}
	\vspace{-10pt}
\end{figure*}

The statistical properties of the dataset are shown in Figure \ref{Fig:statistic}b-d. The dataset's motion types are categorized into translation, rotation, and compound, while its scenes span 8 major categories, including roadside, suburbs, and riverside. The motion intention prompt lengths follow a near-normal distribution, with a mean of 163.9 and median of 160 characters. Based on the video content, text length, and word cloud analysis, the dataset demonstrates diversity and is well-suited for training and testing models to predict future sequence observations.

\section{Learning an Aerial World Model}
\label{sec:method}

An Aerial world model $W$ receives the current state of the world and predicts the future state if an aerial agent performs motion intention. In this work, the current state refers to the current egocentric visual observation $o$. The intention refers to the trajectory, movement, or goal of the aerial agent in 6DoF spaces, expressed in high-level textual form $p$. The future state represents the sequential changes in embodied visual observations, expressed in the form of a video $\hat{v}$. Thus the above process can be expressed as:
\begin{equation}
    \hat{v} = W\left(o,p\right)
\end{equation}

\begin{figure*}[h]
	\centering
    \vspace{-15pt}
	\includegraphics[width = \linewidth]{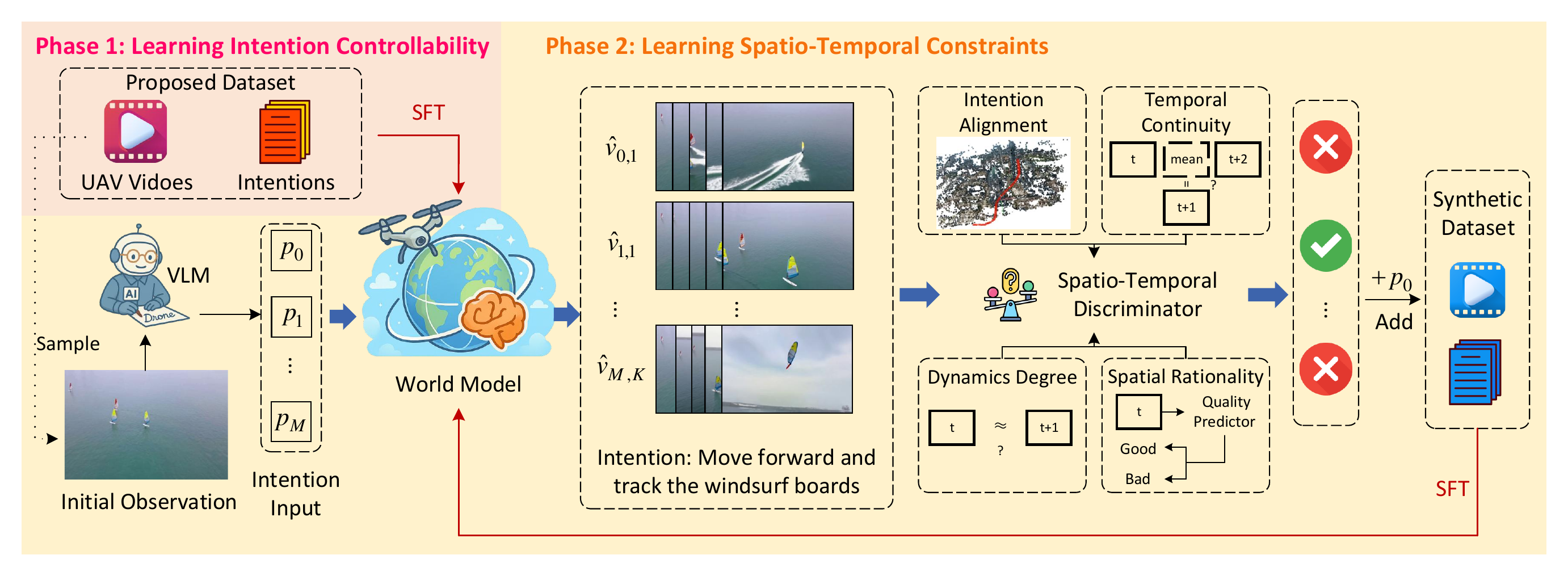}
	\vspace{-25pt}
	\caption{The proposed two-phase training schedule aims to develop an aerial world model that is motion-controllable while adhering to physical spatio-temporal constraints. Phase 1 involves supervised fine-tuning (SFT) on the aerial video-intention pair dataset introduced in Section \ref{sec:dataset}. Phase 2 uses rejection sampling to roll out high-quality samples for iterative SFT. We give an example of this process: The initial frame depicts windsurf boards on the sea, with the drone intending to move forward while keeping them in focus. Among the generated videos, the first is unrealistic as a windsurf board moves like a speedboat, and the last is unreasonable as a board flies into the air. The second video is consistent with real-world physics, with the drone adjusting its gimbal downward to keep the boards in view, making them appear larger in the egocentric perspective.}
	\label{Fig:method}
	\vspace{-10pt}
\end{figure*}

We propose a two-phase training schedule to obtain $W$, as shown in Figure \ref{Fig:method}. First, the foundation model is fine-tuned on the proposed dataset to acquire basic intention controllability. Furthermore, a self-play approach is introduced, where synthetic data is generated and trained based on a spatio-temporal discriminator, ensuring the generated videos adhere to spatio-temporal constraints.

\subsection{Phase 1: Learning Intention Controllability}

Developing a world model based on a pre-trained video generation foundation model can leverage its inherent capability for dynamic modeling in temporal sequences, significantly reducing data and training resource requirements. To enable the video generation foundation model to predict future sequential observations based on the current observation and embodied motion intention, we first perform supervised fine-tuning (SFT). Currently, the foundation model is accustomed to inputs in the form of textual prompts that provide detailed descriptions of the content and specifics of the video to be generated. In this case, the model primarily serves as a tool for visualizing text and images into videos, rather than functioning as the predictive and reasoning world model we aim to develop. For example:
\vspace{-10pt}
\begin{tcolorbox}[boxrule=0mm]
	\textit{As the perspective moves forward, the blue building ahead gradually enlarges in the field of view. In front of it is a road with a continuous stream of cars ...}
\end{tcolorbox}

In contrast, a world model should fully extract the environmental information embedded in the current observation and predict the sequence of observations that would result from an aerial agent executing a series of actions to fulfill its motion intention. For example:
\begin{tcolorbox}[boxrule=0mm]
	\textit{The drone moves forward until it approaches the blue building.}
\end{tcolorbox}

Additionally, the motion of a 6DoF drone involves actions such as lateral translation, vertical movement, rotational adjustments, and gimbal angle changes—scenarios that are underrepresented in the foundation model's pretraining phase. The combination of these actions further increases the search space for future states.

To empower foundation model with the aerial spatial prediction capability, we perform SFT training on the video generation foundation model using the proposed dataset of video and textual intention pairs  \( \mathcal{D} = \{(v_i, p_i)\}_{i=1}^N \), where \( v_i \) represents a video and \( p_i \) is its corresponding textual intention. The fine-tuning process minimizes the reconstruction loss between the predictive outcome \( \hat{v} \) and the ground truth outcome \( v \):
\begin{equation}
    \mathcal{L}_{\text{fine-tune}} = \frac{1}{N} \sum_{i=1}^N \mathcal{L}_{\text{recon}}(W(o_i, p_i), v_i),
\end{equation}
where \( \mathcal{L}_{\text{recon}} \) is a reconstruction loss that measures the similarity between the two videos. \( o_i \) is the initial frame of \( v_i \).

\subsection{Phase 2: Learning Spatio-Temporal Constraints}

After SFT, the generative foundation model can imagine embodied sequential observations resulting from aerial motion intentions. However, for the unique motion scenarios of 6DoF aerial agents, the predicted outcomes remain unstable. Specifically, spatial inconsistencies arise, such as objects with unrealistic shapes (e.g., cars appearing circular) or implausible spatial relationships (e.g., roads floating in the air). Temporally, unnatural deformations occur, such as a pedestrian suddenly splitting into two or buildings continuously twisting. Can the prediction quality of the model be further improved under limited training data?
We propose a self-play training process, which involves generating synthetic data pairs incorporating a spatio-temporal discriminator.

\textbf{a. Motion Intention Generation}. Firstly, we randomly sample a video from the original training dataset and then randomly select a frame from the video. This frame is used as the current observation for the aerial agent $o$. We design a motion prompt $p_{\text{LMM}}$ for the LMM to mimic the intentions present in the training dataset and randomly generate a basic intention $p_0$. Next, the LMM is tasked with expanding the basic intention, generating \(M\) extended intentions $p_1,p_2,...,p_M$, ranging from concise to complex, in the following format: 
\begin{tcolorbox}[boxrule=0mm]
	\textit{\textbf{Basic}: subject + intention \\ \textbf{Extended}: subject + intention \textcolor{blue}{[intention description] + potential outcomes [future observation description]}.}
\end{tcolorbox}
The key insight here is that we aim to obtain multiple linguistic expressions \( \{p_j\}_{j=0}^M \) for the same intention. By leveraging the multimodal understanding and text generation capabilities of the LMM, we can generate coherent and reasonable textual intentions:
\begin{equation}
    \{ {p_j}\} _{j = 0}^M = {\rm{LMM}}(o|{p_{{\rm{LLM}}}}).
\end{equation}

\textbf{b. Video Generation}. For each intention \( p_j \) and the condition observation \( o \), the world model \( W \) generates multiple videos \( \{v_{j,k}\}_{k=1}^K \) using different random seeds \( s_k \):
\begin{equation}
    \hat{v}_{j,k} = W(o, p_j | s_k), \quad j = 0, \dots, M, \quad k = 1, \dots, K.
\end{equation}
This results in a set of candidate videos \( \{\hat{v}_{j,k}\} \) for the similar motion intention of the aerial agent.

\textbf{c. Rejection Sampling}. 
We aim to design a discriminator to identify which video, among multiple inputs with similar motion intentions, better satisfies spatio-temporal constraints. We propose the following four features, which reflect the quality of predicted videos from different perspectives:
\begin{itemize}[leftmargin=*]
    \item \textbf{Intention Alignment $x'$}: This feature evaluates whether the generated video aligns with the intended motion by analyzing differences in implicit trajectories across videos \( \{\hat{v}_{j,k}\} \). First, the 3D environment and trajectory coordinates are reconstructed from each video. An anomaly detection algorithm is then applied to filter out abnormal trajectories. The underlying assumption is that most generated videos adhere to the motion intention, while a small number of divergent motions can be identified. Specifically, we use VGGT~\cite{wang2025vggt} to extract trajectories and isolation forest~\cite{liu2008isolation} to detect anomalous trajectories.
    \item \textbf{Temporal Continuity $x''$}: The states of objects should change continuously over time, without abrupt jumps or discontinuities. In this case, the short-term observation between consecutive frames is approximately linear. Thus, we extract the even-numbered frames from the video and synthesize them by averaging the adjacent odd-numbered frames. The smoothness is then assessed by calculating the Mean Absolute Error between the real and synthesized even-numbered frames~\cite{li2023amt}. 
    \item \textbf{Dynamic Degree $x'''$}: Video foundation models tend to generate results with minimal or no movement~\cite{huang2024vbench}. In aerial motion scenarios, we expect the world model to produce actions with relatively larger motion amplitudes. RAFT~\cite{teed2020raft} is employed to evaluate the dynamics of the generated embodied observations.
    \item \textbf{Spatial Rationality $x''''$}: The generated observations often exhibit chaotic and unrealistic spatial structures, such as distorted buildings or large patches of snow, which are particularly prominent in the final frames. We adapt two pre-trained models, LAION~\cite{laion2022aesthetic} and MUSIQ~\cite{ke2021musiq}, to assess the quality of the final frame, thereby inferring the spatial rationality of the video.
\end{itemize}
The above process can be summarized as a feature extractor \( G \):
\begin{equation}
    \{{x}_{j,k}', {x}_{j,k}'', {x}_{j,k}''', {x}_{j,k}''''\} = G\left( \hat{v}_{j,k} \right)
\end{equation}

We then manually annotated a dataset, selecting videos that best satisfy spatiotemporal constraints, denoted as $\mathcal{D}_{\text{discriminator}}$. We further train a machine learning model $F$ using the four aforementioned features for fitting.

By employing $F$ to output scores for each video, we can obtain a video that aligns with the basic intention \( p_0 \) and satisfies spatio-temporal constraints:
\begin{equation}
    v^* = \arg\max_{v_{j,k}} F\left( G\left(v_{j,k}\right) \right),
\end{equation}

\textbf{d. Synthetic Data Collection}.
The selected video \( v^* \) and its corresponding basic intention prompt \( p_{0} \) form a synthetic data pair \( (v^*, p_{0}) \). This pair is added to the synthetic dataset \( \mathcal{D}_{\text{synthetic}} \).

\begin{equation}
    \mathcal{D}_{\text{synthetic}} \leftarrow \mathcal{D}_{\text{synthetic}} \cup \{(v^*, p_{0})\}.
\end{equation}

\textbf{e. Supervised Fine-Tuning}.
When the size of the synthetic dataset \( \mathcal{D}_{\text{synthetic}} \) reaches the predefined threshold, it is used to further train the world model \( W \). The training objective is similar to the fine-tuning phase, where the reconstruction loss is minimized:

\begin{equation}
    \mathcal{L}_{\text{self-play}} = \frac{1}{|\mathcal{D}_{\text{synthetic}}|} \sum_{(v, p) \in \mathcal{D}_{\text{synthetic}}} \mathcal{L}_{\text{recon}}(W(o, p), v).
\end{equation}


In this process, the critics of the discriminator are utilized to extract high-quality predictions from the world model. These predictions are then enhanced during SFT, while generations that violate spatio-temporal constraints are suppressed, ultimately enabling the prediction of future observations for 6DoF aerial agents.

	\section{Experiments}


\label{sec:experiments}

\subsection{Experimental Setup}

\textbf{Implementation Details.} 
The proposed dataset is randomly divided into training and testing sets with a ratio of 9:1. We build AirScape based on the video generation foundation model CogVideoX-i2v-5B~\cite{yang2024cogvideox}, with main training parameters set as follows: a video resolution of 49$\times$480$\times$720 (frames$\times$height$\times$width), a batch size of 2, gradient accumulation steps of 8, and a total of 10 training epochs. The model was trained on 8 NVIDIA A800-SXM4-40GB GPUs. Additionally, we employed the VLM model Gemini-2.0-Flash~\cite{gemini2023} in the Phase 2 intention generation, which was selected for its superior video understanding capabilities and efficient response speed. The size of $\mathcal{D}_{\text{discriminator}}$ is 500 video groups, each containing 8–16 videos. The machine learning model $F$ is implemented via Random Forest~\cite{breiman2001random}.

\begin{table*}[t]
    \vspace{-7pt}
	\caption{Evaluation results of predictive future sequence observations for 6DoF aerial agents in three-dimensional space.}
    \vspace{-10pt}
	\label{tab::comparison_results}
	\centering
        \setlength{\tabcolsep}{2pt}
	\resizebox{\linewidth}{!}{
		\begin{tabular}{r|ccc|ccc|ccc|ccc}
			\toprule
			\multicolumn{1}{c}{\multirow{2}*{Model}}  &  \multicolumn{3}{c}{Translation} & \multicolumn{3}{c}{Rotation} & \multicolumn{3}{c}{Compound}   & \multicolumn{3}{c}{Average}            \\
			\cmidrule(r){2-4}
			\cmidrule(r){5-7}
			\cmidrule(r){8-10}
                \cmidrule(r){11-13}
			
			~    & FID $\downarrow$     & FVD $\downarrow$  & IAR/\% $\uparrow$ & FID $\downarrow$     & FVD $\downarrow$  & IAR/\% $\uparrow$ & FID $\downarrow$     & FVD $\downarrow$  & IAR/\% $\uparrow$ & FID $\downarrow$     & FVD $\downarrow$  & IAR/\% $\uparrow$ \\ 
			\midrule
            \multicolumn{1}{l|}{\cellcolor[HTML]{F5F5F5}\textit{\textbf{Video Generation Foundation Model}}} & \multicolumn{3}{c|}{\cellcolor[HTML]{F5F5F5}} & \multicolumn{3}{c|}{\cellcolor[HTML]{F5F5F5}} & \multicolumn{3}{c|}{\cellcolor[HTML]{F5F5F5}} & \multicolumn{3}{c}{\cellcolor[HTML]{F5F5F5}}\\
            LTX-Video-2B~\cite{hacohen2024ltx} & 153.42 & 3576.48 &  37.10& 164.52 & 1097.05 &  23.53 & 153.45  & 1002.81 &  19.05& 154.72 & 2600.90 &  26.56\\
            CogVideoX-I2V-5B~\cite{yang2024cogvideox} & 126.24 & 2656.45 & 30.61 & 153.96 & 733.68 & 27.27 &  121.34 & 895.32 & 14.55 & 127.89 & 1947.60 & 24.14 \\
            HunyuanVideo-I2V~\cite{kong2024hunyuanvideo} & 173.03 & 1423.45 &22.41& 216.97 & \bf 614.52 &8.33& 189.77 & \bf 343.73 &35.29& 182.60 & 1043.38 &22.01    \\ 
        Wan2.1-I2V-14B~\cite{wan2025wan} & 150.46 & 2622.07 &  32.35& 183.85 & 1003.68 &  28.57&  165.89 & 1134.29 &  24.00& 158.47 & 2036.52 &  28.31\\ \hline
            \multicolumn{1}{l|}{\cellcolor[HTML]{F5F5F5}\textit{\textbf{World Foundation Model}}} & \multicolumn{3}{c|}{\cellcolor[HTML]{F5F5F5}} & \multicolumn{3}{c|}{\cellcolor[HTML]{F5F5F5}} & \multicolumn{3}{c|}{\cellcolor[HTML]{F5F5F5}} & \multicolumn{3}{c}{\cellcolor[HTML]{F5F5F5}}\\
            Cosmos-Predict2-2B-Video2World~\cite{nvidia_cosmos_predict2} & 236.71 & 2496.94 &  22.81& 255.74 & 942.57 &  33.33& 234.82  & 949.99 &  29.03& 238.43 & 1903.08 &  28.39\\
            Cosmos-Predict1-7B-Video2World~\cite{agarwal2025cosmos} & 142.52 & 2840.73 & 36.21 & 159.60 & 1171.47 & 26.92 &  142.43 & 1263.72 & 32.31 & 144.48 & 2225.45 & 31.81 \\ \hline
            \multicolumn{1}{l|}{\cellcolor[HTML]{F5F5F5}\textit{\textbf{Aerial World Model}}} & \multicolumn{3}{c|}{\cellcolor[HTML]{F5F5F5}} & \multicolumn{3}{c|}{\cellcolor[HTML]{F5F5F5}} & \multicolumn{3}{c|}{\cellcolor[HTML]{F5F5F5}} & \multicolumn{3}{c}{\cellcolor[HTML]{F5F5F5}}\\ 
            \textbf{AirScape} & \textbf{104.07} & \textbf{824.75} & \textbf{84.44} & \textbf{142.67} & 623.53 & \textbf{81.82} & \textbf{114.19} & 468.49 & \textbf{87.27} & \textbf{111.16} & \textbf{701.90} & \textbf{84.51} \\
			\bottomrule
		\end{tabular}}
    \vspace{-7pt}
\end{table*}

\begin{figure*}[h]
	\centering
	\includegraphics[width = 0.95\linewidth]{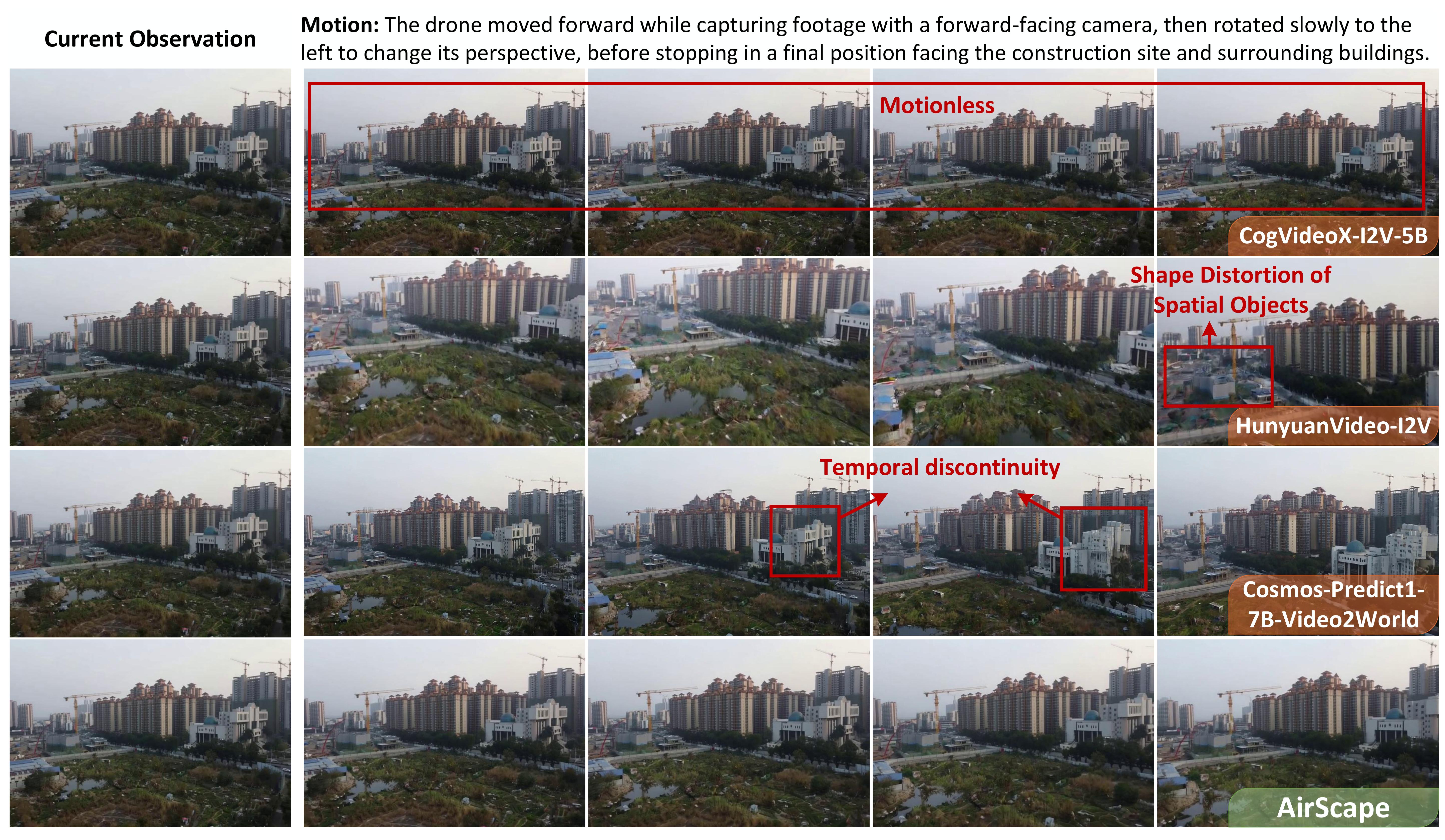}
	\vspace{-13pt}
	\caption{Case analysis of our AirScape and baseline methods, highlighting three common generation issues: limited motion amplitude, shape distortion of spatial objects, and temporal discontinuity.}
	\label{fig::case}
	\vspace{-12pt}
\end{figure*}

\begin{figure*}[h]
	\centering
    \vspace{-11pt}
	\includegraphics[width = 0.95\linewidth]{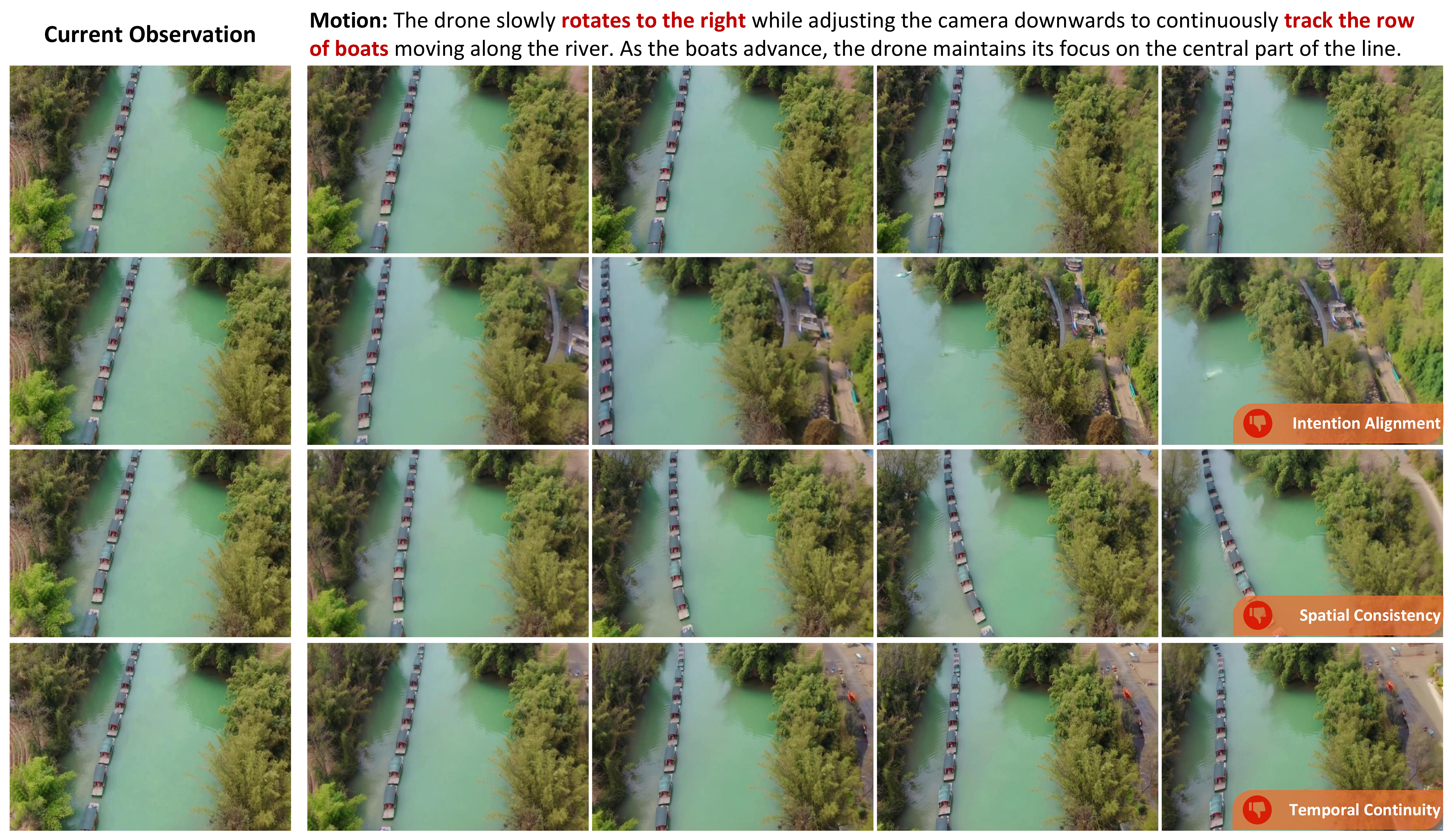}
	\vspace{-13pt}
	\caption{In Phase 2 training, different videos are generated under the same basic intention through rollouts. The spatio-temporal discriminator selects the outcome that best aligns with the intention while satisfying physical spatio-temporal constraints.}
	\label{Fig:self_play}
	\vspace{-10pt}
\end{figure*}

\noindent \textbf{Metrics.} 
We evaluate the quality of the world model's predictive embodied observations from two perspectives: (1) the spatio-temporal distribution differences between the generated videos and the ground truth, and (2) the semantic alignment between the generated videos and the input intention.

\begin{itemize}[leftmargin=*]
	\item Automatic Evaluation: FID~\cite{heusel2017gans} is used to measure the frame-wise distribution differences between the generated videos and the ground truth videos. For FID evaluation, we crop and resize the predicted frames to match the resolution of the ground truth. FVD~\cite{unterthiner2018towards} evaluates the distribution differences in the temporal dimension. For FVD evaluation, all generated videos and ground truth videos are uniformly downsampled to the same number of frames.
	\item Human Evaluation: The counterfactual reasoning capability of the world model requires assessing whether the generated future observations align with the potential outcomes caused by the motion intention. This evaluation involves judging the semantic consistency between the generated videos and the input intention. However, even state-of-the-art video understanding models struggle to accurately capture the semantic relationship between embodied observations and actions~\cite{zhao2025urbanvideobench}. Following the evaluation ideas in~\cite{wang2023videofactory,bar2024lumiere,gao2024vista}, we opt for human evaluation for a more reliable analysis. Specifically, participants are presented with the intention input and the corresponding generated video, and are asked to judge whether they are semantically aligned or not (binary choice). The average intention alignment rate (IAR) is then calculated across the entire test set. For each method, approximately 1.1k generated videos are evaluated, which are randomly and evenly distributed among 9 participants for rating.
\end{itemize}

\noindent \textbf{Baselines.} Due to the absence of world models designed for aerial agents, direct comparisons are not feasible. The most relevant baselines fall into two categories: video generation foundation models and world foundation models. The former includes four popular models: LTX-Video-2B~\cite{hacohen2024ltx}, CogVideoX-I2V-5B~\cite{yang2024cogvideox}, HunyuanVideo-I2V~\cite{kong2024hunyuanvideo}, and Wan2.1-I2V-14B~\cite{wan2025wan}. The latter comprises two different versions of the Cosmos-Predict world models. These models span a parameter range of 2B to 14B.

\subsection{Quantitative Results}

We present the experimental results of model performance in Table~\ref{tab::comparison_results}.
We consider three groups of comparisons, based on which we have the following conclusions.
\begin{itemize}[leftmargin=*]
    \item \textbf{Our proposed AirScape achieves the overall best performance.} Compared with the best-performing baseline models across the three metrics, AirScape outperforms them with average improvements of 15.47\%, 32.73\%, and 52.7\% on FID, FVD, and IAR metrics, respectively. It is worth mentioning in the Rotation group which requries high ability of physical law following, the AirScape's performance gain compared with the best baselines are significant.

    \item \textbf{Outcome prediction of 3D aerial motion is challenging.} We can find that, although HunyuanVideo-I2V achieves the best performance on one metric, FVD, in the Compound group, it has poor performance on FID metric in the same group. This phenomenon also exits for other baselines, indicating that the generation focus and optimization direction of existing baseline models are misaligned. Our AirScape achieves overall stable performance, which further verifies the effectiveness of our design.

    \item \textbf{Model size is not necessarily perfectly correlated with performance.} While larger models generally produce better results, some smaller models can also achieve competitive outcomes. This suggests that there is still room for improving state-of-the-art methods. With carefully curated datasets and advanced training techniques, model performance can be further enhanced.
\end{itemize}



\subsection{Case Analysis}

As shown in Figure \ref{fig::case}, we present an example illustrating the results generated by the baseline models and AirScape. The output from CogVideoX-I2V-5B appears nearly static, indicating a lack of understanding of motion intention. The results from HunyuanVideo-I2V exhibit distortion in the lower-left region of the final frames, which violates spatial physical consistency. For Cosmos-Predict1-7B-Video2World, the white building on the right undergoes abrupt changes in the temporal sequence, failing to maintain temporal continuity. In contrast, the proposed AirScape effectively predicts sequence observations under motion intention while adhering to spatio-temporal constraints.

\subsection{How the Self-Play Works?}

The rolled-out videos exhibit noticeable differences, as shown in Figure \ref{Fig:self_play}. By increasing the number of generations for each input, we aim to ensure that at least one video sample with good prediction quality is generated. The spatio-temporal discriminator evaluates whether videos satisfy the motion intention alignment and spatio-temporal constraints of embodied observations. 
In the second row of videos in Figure \ref{Fig:self_play}, the row of boats disappears from the frame in later stages, violating the intended goal. In the third row, the row of boats undergoes unnatural distortion, breaking spatial consistency. In the fourth row, a hill suddenly appears in the right field of view, disrupting temporal continuity. The first row of videos has the highest rating among the examples. The above process yields a high-quality synthesized dataset.

After further training on the synthesized dataset, the stability of the model's prediction quality improved. As shown in Table~\ref{tab::ablation_study}, the standard deviations of FID and FVD decreased by 2.9\% and 4.8\%, respectively, reducing violations of spatio-temporal constraints in the predictions.

\begin{table}[t]
	\caption{Ablation study of two-phase training schedule.}
    \vspace{-10pt}
	\label{tab::ablation_study}
	\centering
		\begin{tabular}{c|cc|cc}
			\toprule
			\multirow{2}*{Training}  &  \multicolumn{2}{c}{FID $\downarrow$} & \multicolumn{2}{c}{FVD $\downarrow$}  \\
			\cmidrule(r){2-3}
			\cmidrule(r){4-5}
			
			~    & Avg.    & Std.  &  Avg. & Std. \\ 
			\midrule
        After Phase 1 & 110.98 & 722.95 & 59.56 & 1097.18 \\
        After Phase 2 & 111.16 & $701.90^{\color{red}\downarrow 2.9\%}$ & 57.78 & $1044.47^{\color{red}\downarrow 4.8\%}$ \\
			\bottomrule
		\end{tabular}
    \vspace{-18pt}
\end{table}
	\section{Conclusion and Future Work}

This paper introduces the first aerial world model capable of imagining future embodied observational sequences based on motion intentions. We present a dataset comprising 11k video-motion pairs and a two-phase training schedule for foundation models. Experimental results reveal that aerial spatial imagination poses significant challenges to existing models, while our proposed AirScape achieves substantial improvements across all metrics. In the future, we aim to enhance 1) real-time performance, 2) lightweight design, and 3) applicability for assisting decision-making in real-world aerial agent operations.
	
	\section*{Acknowledgments}
	This paper was supported by the Natural Science Foundation of China under Grant 62371269, Shenzhen Science and Technology Plan Project KJZD20240903102700001, Meituan Academy of Robotics Shenzhen, Talent Program of Guangdong Province (2021QN02Z1\\07), National Science and Technology Major Project (2024ZD01NL00\\103) and the Major Key Project of PCL (PCL2025A03).

	\bibliographystyle{ACM-Reference-Format}
	\balance
	\bibliography{sample-base}


\begin{thebibliography}{76}


\ifx \showCODEN    \undefined \def \showCODEN     #1{\unskip}     \fi
\ifx \showISBNx    \undefined \def \showISBNx     #1{\unskip}     \fi
\ifx \showISBNxiii \undefined \def \showISBNxiii  #1{\unskip}     \fi
\ifx \showISSN     \undefined \def \showISSN      #1{\unskip}     \fi
\ifx \showLCCN     \undefined \def \showLCCN      #1{\unskip}     \fi
\ifx \shownote     \undefined \def \shownote      #1{#1}          \fi
\ifx \showarticletitle \undefined \def \showarticletitle #1{#1}   \fi
\ifx \showURL      \undefined \def \showURL       {\relax}        \fi
\providecommand\bibfield[2]{#2}
\providecommand\bibinfo[2]{#2}
\providecommand\natexlab[1]{#1}
\providecommand\showeprint[2][]{arXiv:#2}

\bibitem[Agarwal et~al\mbox{.}(2025)]%
        {agarwal2025cosmos}
\bibfield{author}{\bibinfo{person}{Niket Agarwal}, \bibinfo{person}{Arslan
  Ali}, \bibinfo{person}{Maciej Bala}, \bibinfo{person}{Yogesh Balaji},
  \bibinfo{person}{Erik Barker}, \bibinfo{person}{Tiffany Cai},
  \bibinfo{person}{Prithvijit Chattopadhyay}, \bibinfo{person}{Yongxin Chen},
  \bibinfo{person}{Yin Cui}, \bibinfo{person}{Yifan Ding}, {et~al\mbox{.}}}
  \bibinfo{year}{2025}\natexlab{}.
\newblock \showarticletitle{Cosmos world foundation model platform for physical
  ai}.
\newblock \bibinfo{journal}{\emph{arXiv preprint arXiv:2501.03575}}
  (\bibinfo{year}{2025}).
\newblock


\bibitem[AI(2023)]%
        {gemini2023}
\bibfield{author}{\bibinfo{person}{Google AI}.}
  \bibinfo{year}{2023}\natexlab{}.
\newblock \bibinfo{title}{Gemini-2.0-Flash}.
\newblock \bibinfo{howpublished}{\url{https://ai.google.dev/}}.
\newblock
\newblock
\shownote{Accessed: 2025-05-30}.


\bibitem[An et~al\mbox{.}(2024)]%
        {an2024etpnav}
\bibfield{author}{\bibinfo{person}{Dong An}, \bibinfo{person}{Hanqing Wang},
  \bibinfo{person}{Wenguan Wang}, \bibinfo{person}{Zun Wang},
  \bibinfo{person}{Yan Huang}, \bibinfo{person}{Keji He}, {and}
  \bibinfo{person}{Liang Wang}.} \bibinfo{year}{2024}\natexlab{}.
\newblock \showarticletitle{Etpnav: Evolving topological planning for
  vision-language navigation in continuous environments}.
\newblock \bibinfo{journal}{\emph{IEEE Transactions on Pattern Analysis and
  Machine Intelligence}} (\bibinfo{year}{2024}).
\newblock


\bibitem[Bar-Tal et~al\mbox{.}(2024)]%
        {bar2024lumiere}
\bibfield{author}{\bibinfo{person}{Omer Bar-Tal}, \bibinfo{person}{Hila
  Chefer}, \bibinfo{person}{Omer Tov}, \bibinfo{person}{Charles Herrmann},
  \bibinfo{person}{Roni Paiss}, \bibinfo{person}{Shiran Zada},
  \bibinfo{person}{Ariel Ephrat}, \bibinfo{person}{Junhwa Hur},
  \bibinfo{person}{Guanghui Liu}, \bibinfo{person}{Amit Raj}, {et~al\mbox{.}}}
  \bibinfo{year}{2024}\natexlab{}.
\newblock \showarticletitle{Lumiere: A space-time diffusion model for video
  generation}. In \bibinfo{booktitle}{\emph{SIGGRAPH Asia 2024 Conference
  Papers}}. \bibinfo{pages}{1--11}.
\newblock


\bibitem[Blattmann et~al\mbox{.}(2023)]%
        {blattmann2023stable}
\bibfield{author}{\bibinfo{person}{Andreas Blattmann}, \bibinfo{person}{Tim
  Dockhorn}, \bibinfo{person}{Sumith Kulal}, \bibinfo{person}{Daniel
  Mendelevitch}, \bibinfo{person}{Maciej Kilian}, \bibinfo{person}{Dominik
  Lorenz}, \bibinfo{person}{Yam Levi}, \bibinfo{person}{Zion English},
  \bibinfo{person}{Vikram Voleti}, \bibinfo{person}{Adam Letts},
  {et~al\mbox{.}}} \bibinfo{year}{2023}\natexlab{}.
\newblock \showarticletitle{Stable video diffusion: Scaling latent video
  diffusion models to large datasets}.
\newblock \bibinfo{journal}{\emph{arXiv preprint arXiv:2311.15127}}
  (\bibinfo{year}{2023}).
\newblock


\bibitem[Breiman(2001)]%
        {breiman2001random}
\bibfield{author}{\bibinfo{person}{Leo Breiman}.}
  \bibinfo{year}{2001}\natexlab{}.
\newblock \showarticletitle{Random forests}.
\newblock \bibinfo{journal}{\emph{Machine learning}} \bibinfo{volume}{45},
  \bibinfo{number}{1} (\bibinfo{year}{2001}), \bibinfo{pages}{5--32}.
\newblock


\bibitem[Bruce et~al\mbox{.}(2024)]%
        {bruce2024genie}
\bibfield{author}{\bibinfo{person}{Jake Bruce}, \bibinfo{person}{Michael~D
  Dennis}, \bibinfo{person}{Ashley Edwards}, \bibinfo{person}{Jack
  Parker-Holder}, \bibinfo{person}{Yuge Shi}, \bibinfo{person}{Edward Hughes},
  \bibinfo{person}{Matthew Lai}, \bibinfo{person}{Aditi Mavalankar},
  \bibinfo{person}{Richie Steigerwald}, \bibinfo{person}{Chris Apps},
  {et~al\mbox{.}}} \bibinfo{year}{2024}\natexlab{}.
\newblock \showarticletitle{Genie: Generative interactive environments}. In
  \bibinfo{booktitle}{\emph{Forty-first International Conference on Machine
  Learning}}.
\newblock


\bibitem[Caesar et~al\mbox{.}(2020)]%
        {caesar2020nuscenes}
\bibfield{author}{\bibinfo{person}{Holger Caesar}, \bibinfo{person}{Varun
  Bankiti}, \bibinfo{person}{Alex~H Lang}, \bibinfo{person}{Sourabh Vora},
  \bibinfo{person}{Venice~Erin Liong}, \bibinfo{person}{Qiang Xu},
  \bibinfo{person}{Anush Krishnan}, \bibinfo{person}{Yu Pan},
  \bibinfo{person}{Giancarlo Baldan}, {and} \bibinfo{person}{Oscar Beijbom}.}
  \bibinfo{year}{2020}\natexlab{}.
\newblock \showarticletitle{nuscenes: A multimodal dataset for autonomous
  driving}. In \bibinfo{booktitle}{\emph{Proceedings of the IEEE/CVF conference
  on computer vision and pattern recognition}}. \bibinfo{pages}{11621--11631}.
\newblock


\bibitem[Cao et~al\mbox{.}(2024)]%
        {cao2024survey}
\bibfield{author}{\bibinfo{person}{Hanqun Cao}, \bibinfo{person}{Cheng Tan},
  \bibinfo{person}{Zhangyang Gao}, \bibinfo{person}{Yilun Xu},
  \bibinfo{person}{Guangyong Chen}, \bibinfo{person}{Pheng-Ann Heng}, {and}
  \bibinfo{person}{Stan~Z Li}.} \bibinfo{year}{2024}\natexlab{}.
\newblock \showarticletitle{A survey on generative diffusion models}.
\newblock \bibinfo{journal}{\emph{IEEE Transactions on Knowledge and Data
  Engineering}} (\bibinfo{year}{2024}).
\newblock


\bibitem[Chen et~al\mbox{.}(2020)]%
        {chen2020generative}
\bibfield{author}{\bibinfo{person}{Mark Chen}, \bibinfo{person}{Alec Radford},
  \bibinfo{person}{Rewon Child}, \bibinfo{person}{Jeffrey Wu},
  \bibinfo{person}{Heewoo Jun}, \bibinfo{person}{David Luan}, {and}
  \bibinfo{person}{Ilya Sutskever}.} \bibinfo{year}{2020}\natexlab{}.
\newblock \showarticletitle{Generative pretraining from pixels}. In
  \bibinfo{booktitle}{\emph{International conference on machine learning}}.
  PMLR, \bibinfo{pages}{1691--1703}.
\newblock


\bibitem[Chen et~al\mbox{.}(2024a)]%
        {chen2024ddl}
\bibfield{author}{\bibinfo{person}{Xuecheng Chen}, \bibinfo{person}{Haoyang
  Wang}, \bibinfo{person}{Yuhan Cheng}, \bibinfo{person}{Haohao Fu},
  \bibinfo{person}{Yuxuan Liu}, \bibinfo{person}{Fan Dang},
  \bibinfo{person}{Yunhao Liu}, \bibinfo{person}{Jinqiang Cui}, {and}
  \bibinfo{person}{Xinlei Chen}.} \bibinfo{year}{2024}\natexlab{a}.
\newblock \showarticletitle{Ddl: Empowering delivery drones with large-scale
  urban sensing capability}.
\newblock \bibinfo{journal}{\emph{IEEE Journal of Selected Topics in Signal
  Processing}} (\bibinfo{year}{2024}).
\newblock


\bibitem[Chen et~al\mbox{.}(2024b)]%
        {chen2024soscheduler}
\bibfield{author}{\bibinfo{person}{Xuecheng Chen}, \bibinfo{person}{Zijian
  Xiao}, \bibinfo{person}{Yuhan Cheng}, \bibinfo{person}{Chen-Chun Hsia},
  \bibinfo{person}{Haoyang Wang}, \bibinfo{person}{Jingao Xu},
  \bibinfo{person}{Susu Xu}, \bibinfo{person}{Fan Dang},
  \bibinfo{person}{Xiao-Ping Zhang}, \bibinfo{person}{Yunhao Liu},
  {et~al\mbox{.}}} \bibinfo{year}{2024}\natexlab{b}.
\newblock \showarticletitle{Soscheduler: Toward proactive and adaptive wildfire
  suppression via multi-uav collaborative scheduling}.
\newblock \bibinfo{journal}{\emph{IEEE Internet of Things Journal}}
  \bibinfo{volume}{11}, \bibinfo{number}{14} (\bibinfo{year}{2024}),
  \bibinfo{pages}{24858--24871}.
\newblock


\bibitem[Cheng et~al\mbox{.}(2024)]%
        {cheng2024videollama}
\bibfield{author}{\bibinfo{person}{Zesen Cheng}, \bibinfo{person}{Sicong Leng},
  \bibinfo{person}{Hang Zhang}, \bibinfo{person}{Yifei Xin},
  \bibinfo{person}{Xin Li}, \bibinfo{person}{Guanzheng Chen},
  \bibinfo{person}{Yongxin Zhu}, \bibinfo{person}{Wenqi Zhang},
  \bibinfo{person}{Ziyang Luo}, \bibinfo{person}{Deli Zhao}, {et~al\mbox{.}}}
  \bibinfo{year}{2024}\natexlab{}.
\newblock \showarticletitle{Videollama 2: Advancing spatial-temporal modeling
  and audio understanding in video-llms}.
\newblock \bibinfo{journal}{\emph{arXiv preprint arXiv:2406.07476}}
  (\bibinfo{year}{2024}).
\newblock


\bibitem[Chu et~al\mbox{.}(2020)]%
        {chu2020learning}
\bibfield{author}{\bibinfo{person}{Mengyu Chu}, \bibinfo{person}{You Xie},
  \bibinfo{person}{Jonas Mayer}, \bibinfo{person}{Laura Leal-Taix{\'e}}, {and}
  \bibinfo{person}{Nils Thuerey}.} \bibinfo{year}{2020}\natexlab{}.
\newblock \showarticletitle{Learning temporal coherence via self-supervision
  for GAN-based video generation}.
\newblock \bibinfo{journal}{\emph{ACM Transactions on Graphics (TOG)}}
  \bibinfo{volume}{39}, \bibinfo{number}{4} (\bibinfo{year}{2020}),
  \bibinfo{pages}{75--1}.
\newblock


\bibitem[Croitoru et~al\mbox{.}(2023)]%
        {croitoru2023diffusion}
\bibfield{author}{\bibinfo{person}{Florinel-Alin Croitoru},
  \bibinfo{person}{Vlad Hondru}, \bibinfo{person}{Radu~Tudor Ionescu}, {and}
  \bibinfo{person}{Mubarak Shah}.} \bibinfo{year}{2023}\natexlab{}.
\newblock \showarticletitle{Diffusion models in vision: A survey}.
\newblock \bibinfo{journal}{\emph{IEEE Transactions on Pattern Analysis and
  Machine Intelligence}} \bibinfo{volume}{45}, \bibinfo{number}{9}
  (\bibinfo{year}{2023}), \bibinfo{pages}{10850--10869}.
\newblock


\bibitem[Dasari et~al\mbox{.}(2019)]%
        {dasari2019robonet}
\bibfield{author}{\bibinfo{person}{Sudeep Dasari}, \bibinfo{person}{Frederik
  Ebert}, \bibinfo{person}{Stephen Tian}, \bibinfo{person}{Suraj Nair},
  \bibinfo{person}{Bernadette Bucher}, \bibinfo{person}{Karl Schmeckpeper},
  \bibinfo{person}{Siddharth Singh}, \bibinfo{person}{Sergey Levine}, {and}
  \bibinfo{person}{Chelsea Finn}.} \bibinfo{year}{2019}\natexlab{}.
\newblock \showarticletitle{Robonet: Large-scale multi-robot learning}.
\newblock \bibinfo{journal}{\emph{arXiv preprint arXiv:1910.11215}}
  (\bibinfo{year}{2019}).
\newblock


\bibitem[Ding et~al\mbox{.}(2024)]%
        {ding2024understanding}
\bibfield{author}{\bibinfo{person}{Jingtao Ding}, \bibinfo{person}{Yunke
  Zhang}, \bibinfo{person}{Yu Shang}, \bibinfo{person}{Yuheng Zhang},
  \bibinfo{person}{Zefang Zong}, \bibinfo{person}{Jie Feng},
  \bibinfo{person}{Yuan Yuan}, \bibinfo{person}{Hongyuan Su},
  \bibinfo{person}{Nian Li}, \bibinfo{person}{Nicholas Sukiennik},
  {et~al\mbox{.}}} \bibinfo{year}{2024}\natexlab{}.
\newblock \showarticletitle{Understanding world or predicting future? a
  comprehensive survey of world models}.
\newblock \bibinfo{journal}{\emph{Comput. Surveys}} (\bibinfo{year}{2024}).
\newblock


\bibitem[Gao et~al\mbox{.}(2024c)]%
        {gao2024embodiedcity}
\bibfield{author}{\bibinfo{person}{Chen Gao}, \bibinfo{person}{Baining Zhao},
  \bibinfo{person}{Weichen Zhang}, \bibinfo{person}{Jinzhu Mao},
  \bibinfo{person}{Jun Zhang}, \bibinfo{person}{Zhiheng Zheng},
  \bibinfo{person}{Fanhang Man}, \bibinfo{person}{Jianjie Fang},
  \bibinfo{person}{Zile Zhou}, \bibinfo{person}{Jinqiang Cui}, {et~al\mbox{.}}}
  \bibinfo{year}{2024}\natexlab{c}.
\newblock \showarticletitle{EmbodiedCity: A Benchmark Platform for Embodied
  Agent in Real-world City Environment}.
\newblock \bibinfo{journal}{\emph{arXiv preprint arXiv:2410.09604}}
  (\bibinfo{year}{2024}).
\newblock


\bibitem[Gao et~al\mbox{.}(2024a)]%
        {gao2024magicdrive3d}
\bibfield{author}{\bibinfo{person}{Ruiyuan Gao}, \bibinfo{person}{Kai Chen},
  \bibinfo{person}{Zhihao Li}, \bibinfo{person}{Lanqing Hong},
  \bibinfo{person}{Zhenguo Li}, {and} \bibinfo{person}{Qiang Xu}.}
  \bibinfo{year}{2024}\natexlab{a}.
\newblock \showarticletitle{Magicdrive3d: Controllable 3d generation for
  any-view rendering in street scenes}.
\newblock \bibinfo{journal}{\emph{arXiv preprint arXiv:2405.14475}}
  (\bibinfo{year}{2024}).
\newblock


\bibitem[Gao et~al\mbox{.}(2023)]%
        {gao2023magicdrive}
\bibfield{author}{\bibinfo{person}{Ruiyuan Gao}, \bibinfo{person}{Kai Chen},
  \bibinfo{person}{Enze Xie}, \bibinfo{person}{Lanqing Hong},
  \bibinfo{person}{Zhenguo Li}, \bibinfo{person}{Dit-Yan Yeung}, {and}
  \bibinfo{person}{Qiang Xu}.} \bibinfo{year}{2023}\natexlab{}.
\newblock \showarticletitle{Magicdrive: Street view generation with diverse 3d
  geometry control}.
\newblock \bibinfo{journal}{\emph{arXiv preprint arXiv:2310.02601}}
  (\bibinfo{year}{2023}).
\newblock


\bibitem[Gao et~al\mbox{.}(2024b)]%
        {gao2024vista}
\bibfield{author}{\bibinfo{person}{Shenyuan Gao}, \bibinfo{person}{Jiazhi
  Yang}, \bibinfo{person}{Li Chen}, \bibinfo{person}{Kashyap Chitta},
  \bibinfo{person}{Yihang Qiu}, \bibinfo{person}{Andreas Geiger},
  \bibinfo{person}{Jun Zhang}, {and} \bibinfo{person}{Hongyang Li}.}
  \bibinfo{year}{2024}\natexlab{b}.
\newblock \showarticletitle{Vista: A generalizable driving world model with
  high fidelity and versatile controllability}.
\newblock \bibinfo{journal}{\emph{arXiv preprint arXiv:2405.17398}}
  (\bibinfo{year}{2024}).
\newblock


\bibitem[Gu et~al\mbox{.}(2024)]%
        {gu2024advancing}
\bibfield{author}{\bibinfo{person}{Xinyang Gu}, \bibinfo{person}{Yen-Jen Wang},
  \bibinfo{person}{Xiang Zhu}, \bibinfo{person}{Chengming Shi},
  \bibinfo{person}{Yanjiang Guo}, \bibinfo{person}{Yichen Liu}, {and}
  \bibinfo{person}{Jianyu Chen}.} \bibinfo{year}{2024}\natexlab{}.
\newblock \showarticletitle{Advancing humanoid locomotion: Mastering
  challenging terrains with denoising world model learning}.
\newblock \bibinfo{journal}{\emph{arXiv preprint arXiv:2408.14472}}
  (\bibinfo{year}{2024}).
\newblock


\bibitem[Guan et~al\mbox{.}(2024)]%
        {guan2024world}
\bibfield{author}{\bibinfo{person}{Yanchen Guan}, \bibinfo{person}{Haicheng
  Liao}, \bibinfo{person}{Zhenning Li}, \bibinfo{person}{Jia Hu},
  \bibinfo{person}{Runze Yuan}, \bibinfo{person}{Guohui Zhang}, {and}
  \bibinfo{person}{Chengzhong Xu}.} \bibinfo{year}{2024}\natexlab{}.
\newblock \showarticletitle{World models for autonomous driving: An initial
  survey}.
\newblock \bibinfo{journal}{\emph{IEEE Transactions on Intelligent Vehicles}}
  (\bibinfo{year}{2024}).
\newblock


\bibitem[Gupta et~al\mbox{.}(2021)]%
        {gupta2021embodied}
\bibfield{author}{\bibinfo{person}{Agrim Gupta}, \bibinfo{person}{Silvio
  Savarese}, \bibinfo{person}{Surya Ganguli}, {and} \bibinfo{person}{Li
  Fei-Fei}.} \bibinfo{year}{2021}\natexlab{}.
\newblock \showarticletitle{Embodied intelligence via learning and evolution}.
\newblock \bibinfo{journal}{\emph{Nature communications}} \bibinfo{volume}{12},
  \bibinfo{number}{1} (\bibinfo{year}{2021}), \bibinfo{pages}{5721}.
\newblock


\bibitem[Ha and Schmidhuber(2018)]%
        {ha2018world}
\bibfield{author}{\bibinfo{person}{David Ha} {and} \bibinfo{person}{J{\"u}rgen
  Schmidhuber}.} \bibinfo{year}{2018}\natexlab{}.
\newblock \showarticletitle{World models}.
\newblock \bibinfo{journal}{\emph{arXiv preprint arXiv:1803.10122}}
  (\bibinfo{year}{2018}).
\newblock


\bibitem[HaCohen et~al\mbox{.}(2024)]%
        {hacohen2024ltx}
\bibfield{author}{\bibinfo{person}{Yoav HaCohen}, \bibinfo{person}{Nisan
  Chiprut}, \bibinfo{person}{Benny Brazowski}, \bibinfo{person}{Daniel Shalem},
  \bibinfo{person}{Dudu Moshe}, \bibinfo{person}{Eitan Richardson},
  \bibinfo{person}{Eran Levin}, \bibinfo{person}{Guy Shiran},
  \bibinfo{person}{Nir Zabari}, \bibinfo{person}{Ori Gordon}, {et~al\mbox{.}}}
  \bibinfo{year}{2024}\natexlab{}.
\newblock \showarticletitle{Ltx-video: Realtime video latent diffusion}.
\newblock \bibinfo{journal}{\emph{arXiv preprint arXiv:2501.00103}}
  (\bibinfo{year}{2024}).
\newblock


\bibitem[Heusel et~al\mbox{.}(2017)]%
        {heusel2017gans}
\bibfield{author}{\bibinfo{person}{Martin Heusel}, \bibinfo{person}{Hubert
  Ramsauer}, \bibinfo{person}{Thomas Unterthiner}, \bibinfo{person}{Bernhard
  Nessler}, {and} \bibinfo{person}{Sepp Hochreiter}.}
  \bibinfo{year}{2017}\natexlab{}.
\newblock \showarticletitle{Gans trained by a two time-scale update rule
  converge to a local nash equilibrium}.
\newblock \bibinfo{journal}{\emph{Advances in neural information processing
  systems}}  \bibinfo{volume}{30} (\bibinfo{year}{2017}).
\newblock


\bibitem[Ho et~al\mbox{.}(2020)]%
        {NEURIPS2020_4c5bcfec}
\bibfield{author}{\bibinfo{person}{Jonathan Ho}, \bibinfo{person}{Ajay Jain},
  {and} \bibinfo{person}{Pieter Abbeel}.} \bibinfo{year}{2020}\natexlab{}.
\newblock \showarticletitle{Denoising Diffusion Probabilistic Models}. In
  \bibinfo{booktitle}{\emph{Advances in Neural Information Processing
  Systems}}, \bibfield{editor}{\bibinfo{person}{H.~Larochelle},
  \bibinfo{person}{M.~Ranzato}, \bibinfo{person}{R.~Hadsell},
  \bibinfo{person}{M.F. Balcan}, {and} \bibinfo{person}{H.~Lin}} (Eds.),
  Vol.~\bibinfo{volume}{33}. \bibinfo{publisher}{Curran Associates, Inc.},
  \bibinfo{pages}{6840--6851}.
\newblock
\urldef\tempurl%
\url{https://proceedings.neurips.cc/paper_files/paper/2020/file/4c5bcfec8584af0d967f1ab10179ca4b-Paper.pdf}
\showURL{%
\tempurl}


\bibitem[Ho et~al\mbox{.}(2022)]%
        {ho2022video}
\bibfield{author}{\bibinfo{person}{Jonathan Ho}, \bibinfo{person}{Tim
  Salimans}, \bibinfo{person}{Alexey Gritsenko}, \bibinfo{person}{William
  Chan}, \bibinfo{person}{Mohammad Norouzi}, {and} \bibinfo{person}{David~J
  Fleet}.} \bibinfo{year}{2022}\natexlab{}.
\newblock \showarticletitle{Video diffusion models}.
\newblock \bibinfo{journal}{\emph{Advances in neural information processing
  systems}}  \bibinfo{volume}{35} (\bibinfo{year}{2022}),
  \bibinfo{pages}{8633--8646}.
\newblock


\bibitem[Hong et~al\mbox{.}(2022)]%
        {hong2022cogvideo}
\bibfield{author}{\bibinfo{person}{Wenyi Hong}, \bibinfo{person}{Ming Ding},
  \bibinfo{person}{Wendi Zheng}, \bibinfo{person}{Xinghan Liu}, {and}
  \bibinfo{person}{Jie Tang}.} \bibinfo{year}{2022}\natexlab{}.
\newblock \showarticletitle{Cogvideo: Large-scale pretraining for text-to-video
  generation via transformers}.
\newblock \bibinfo{journal}{\emph{arXiv preprint arXiv:2205.15868}}
  (\bibinfo{year}{2022}).
\newblock


\bibitem[Hu et~al\mbox{.}(2022)]%
        {hu2022make}
\bibfield{author}{\bibinfo{person}{Yaosi Hu}, \bibinfo{person}{Chong Luo},
  {and} \bibinfo{person}{Zhenzhong Chen}.} \bibinfo{year}{2022}\natexlab{}.
\newblock \showarticletitle{Make it move: controllable image-to-video
  generation with text descriptions}. In \bibinfo{booktitle}{\emph{Proceedings
  of the IEEE/CVF Conference on Computer Vision and Pattern Recognition}}.
  \bibinfo{pages}{18219--18228}.
\newblock


\bibitem[Huang et~al\mbox{.}(2024)]%
        {huang2024vbench}
\bibfield{author}{\bibinfo{person}{Ziqi Huang}, \bibinfo{person}{Yinan He},
  \bibinfo{person}{Jiashuo Yu}, \bibinfo{person}{Fan Zhang},
  \bibinfo{person}{Chenyang Si}, \bibinfo{person}{Yuming Jiang},
  \bibinfo{person}{Yuanhan Zhang}, \bibinfo{person}{Tianxing Wu},
  \bibinfo{person}{Qingyang Jin}, \bibinfo{person}{Nattapol Chanpaisit},
  {et~al\mbox{.}}} \bibinfo{year}{2024}\natexlab{}.
\newblock \showarticletitle{Vbench: Comprehensive benchmark suite for video
  generative models}. In \bibinfo{booktitle}{\emph{Proceedings of the IEEE/CVF
  Conference on Computer Vision and Pattern Recognition}}.
  \bibinfo{pages}{21807--21818}.
\newblock


\bibitem[Jiang et~al\mbox{.}(2024)]%
        {jiang2024videobooth}
\bibfield{author}{\bibinfo{person}{Yuming Jiang}, \bibinfo{person}{Tianxing
  Wu}, \bibinfo{person}{Shuai Yang}, \bibinfo{person}{Chenyang Si},
  \bibinfo{person}{Dahua Lin}, \bibinfo{person}{Yu Qiao},
  \bibinfo{person}{Chen~Change Loy}, {and} \bibinfo{person}{Ziwei Liu}.}
  \bibinfo{year}{2024}\natexlab{}.
\newblock \showarticletitle{Videobooth: Diffusion-based video generation with
  image prompts}. In \bibinfo{booktitle}{\emph{Proceedings of the IEEE/CVF
  Conference on Computer Vision and Pattern Recognition}}.
  \bibinfo{pages}{6689--6700}.
\newblock


\bibitem[Ke et~al\mbox{.}(2021)]%
        {ke2021musiq}
\bibfield{author}{\bibinfo{person}{Junjie Ke}, \bibinfo{person}{Qifei Wang},
  \bibinfo{person}{Yilin Wang}, \bibinfo{person}{Peyman Milanfar}, {and}
  \bibinfo{person}{Feng Yang}.} \bibinfo{year}{2021}\natexlab{}.
\newblock \showarticletitle{Musiq: Multi-scale image quality transformer}. In
  \bibinfo{booktitle}{\emph{Proceedings of the IEEE/CVF international
  conference on computer vision}}. \bibinfo{pages}{5148--5157}.
\newblock


\bibitem[Kondratyuk et~al\mbox{.}(2023)]%
        {kondratyuk2023videopoet}
\bibfield{author}{\bibinfo{person}{Dan Kondratyuk}, \bibinfo{person}{Lijun Yu},
  \bibinfo{person}{Xiuye Gu}, \bibinfo{person}{Jos{\'e} Lezama},
  \bibinfo{person}{Jonathan Huang}, \bibinfo{person}{Grant Schindler},
  \bibinfo{person}{Rachel Hornung}, \bibinfo{person}{Vighnesh Birodkar},
  \bibinfo{person}{Jimmy Yan}, \bibinfo{person}{Ming-Chang Chiu},
  {et~al\mbox{.}}} \bibinfo{year}{2023}\natexlab{}.
\newblock \showarticletitle{Videopoet: A large language model for zero-shot
  video generation}.
\newblock \bibinfo{journal}{\emph{arXiv preprint arXiv:2312.14125}}
  (\bibinfo{year}{2023}).
\newblock


\bibitem[Kong et~al\mbox{.}(2024)]%
        {kong2024hunyuanvideo}
\bibfield{author}{\bibinfo{person}{Weijie Kong}, \bibinfo{person}{Qi Tian},
  \bibinfo{person}{Zijian Zhang}, \bibinfo{person}{Rox Min},
  \bibinfo{person}{Zuozhuo Dai}, \bibinfo{person}{Jin Zhou},
  \bibinfo{person}{Jiangfeng Xiong}, \bibinfo{person}{Xin Li},
  \bibinfo{person}{Bo Wu}, \bibinfo{person}{Jianwei Zhang}, {et~al\mbox{.}}}
  \bibinfo{year}{2024}\natexlab{}.
\newblock \showarticletitle{Hunyuanvideo: A systematic framework for large
  video generative models}.
\newblock \bibinfo{journal}{\emph{arXiv preprint arXiv:2412.03603}}
  (\bibinfo{year}{2024}).
\newblock


\bibitem[LAION-AI(2022)]%
        {laion2022aesthetic}
\bibfield{author}{\bibinfo{person}{LAION-AI}.} \bibinfo{year}{2022}\natexlab{}.
\newblock \bibinfo{title}{aesthetic-predictor}.
\newblock
  \bibinfo{howpublished}{\url{https://github.com/LAION-AI/aesthetic-predictor}}.
\newblock


\bibitem[LeCun(2022)]%
        {lecun2022path}
\bibfield{author}{\bibinfo{person}{Yann LeCun}.}
  \bibinfo{year}{2022}\natexlab{}.
\newblock \showarticletitle{A path towards autonomous machine intelligence
  version 0.9. 2, 2022-06-27}.
\newblock \bibinfo{journal}{\emph{Open Review}} \bibinfo{volume}{62},
  \bibinfo{number}{1} (\bibinfo{year}{2022}), \bibinfo{pages}{1--62}.
\newblock


\bibitem[Li et~al\mbox{.}(2025)]%
        {li2025robotic}
\bibfield{author}{\bibinfo{person}{Chenhao Li}, \bibinfo{person}{Andreas
  Krause}, {and} \bibinfo{person}{Marco Hutter}.}
  \bibinfo{year}{2025}\natexlab{}.
\newblock \showarticletitle{Robotic world model: A neural network simulator for
  robust policy optimization in robotics}.
\newblock \bibinfo{journal}{\emph{arXiv preprint arXiv:2501.10100}}
  (\bibinfo{year}{2025}).
\newblock


\bibitem[Li et~al\mbox{.}(2023)]%
        {li2023amt}
\bibfield{author}{\bibinfo{person}{Zhen Li}, \bibinfo{person}{Zuo-Liang Zhu},
  \bibinfo{person}{Ling-Hao Han}, \bibinfo{person}{Qibin Hou},
  \bibinfo{person}{Chun-Le Guo}, {and} \bibinfo{person}{Ming-Ming Cheng}.}
  \bibinfo{year}{2023}\natexlab{}.
\newblock \showarticletitle{Amt: All-pairs multi-field transforms for efficient
  frame interpolation}. In \bibinfo{booktitle}{\emph{Proceedings of the
  IEEE/CVF Conference on Computer Vision and Pattern Recognition}}.
  \bibinfo{pages}{9801--9810}.
\newblock


\bibitem[Liu et~al\mbox{.}(2008)]%
        {liu2008isolation}
\bibfield{author}{\bibinfo{person}{Fei~Tony Liu}, \bibinfo{person}{Kai~Ming
  Ting}, {and} \bibinfo{person}{Zhi-Hua Zhou}.}
  \bibinfo{year}{2008}\natexlab{}.
\newblock \showarticletitle{Isolation forest}. In
  \bibinfo{booktitle}{\emph{2008 eighth ieee international conference on data
  mining}}. IEEE, \bibinfo{pages}{413--422}.
\newblock


\bibitem[Liu et~al\mbox{.}(2024a)]%
        {liu2024mardini}
\bibfield{author}{\bibinfo{person}{Haozhe Liu}, \bibinfo{person}{Shikun Liu},
  \bibinfo{person}{Zijian Zhou}, \bibinfo{person}{Mengmeng Xu},
  \bibinfo{person}{Yanping Xie}, \bibinfo{person}{Xiao Han},
  \bibinfo{person}{Juan~C P{\'e}rez}, \bibinfo{person}{Ding Liu},
  \bibinfo{person}{Kumara Kahatapitiya}, \bibinfo{person}{Menglin Jia},
  {et~al\mbox{.}}} \bibinfo{year}{2024}\natexlab{a}.
\newblock \showarticletitle{Mardini: Masked autoregressive diffusion for video
  generation at scale}.
\newblock \bibinfo{journal}{\emph{arXiv preprint arXiv:2410.20280}}
  (\bibinfo{year}{2024}).
\newblock


\bibitem[Liu et~al\mbox{.}(2024b)]%
        {liu2024mobiair}
\bibfield{author}{\bibinfo{person}{Yuxuan Liu}, \bibinfo{person}{Haoyang Wang},
  \bibinfo{person}{Fanhang Man}, \bibinfo{person}{Jingao Xu},
  \bibinfo{person}{Fan Dang}, \bibinfo{person}{Yunhao Liu},
  \bibinfo{person}{Xiao-Ping Zhang}, {and} \bibinfo{person}{Xinlei Chen}.}
  \bibinfo{year}{2024}\natexlab{b}.
\newblock \showarticletitle{Mobiair: Unleashing sensor mobility for city-scale
  and fine-grained air-quality monitoring with Airbert}. In
  \bibinfo{booktitle}{\emph{Proceedings of the 22nd Annual International
  Conference on Mobile Systems, Applications and Services}}.
  \bibinfo{pages}{223--236}.
\newblock


\bibitem[Liu et~al\mbox{.}(2025)]%
        {liu2025meal}
\bibfield{author}{\bibinfo{person}{Zhishuo Liu}, \bibinfo{person}{Xingquan
  Zuo}, \bibinfo{person}{Mengchu Zhou}, \bibinfo{person}{Bin Jia}, {and}
  \bibinfo{person}{Chongyang Xin}.} \bibinfo{year}{2025}\natexlab{}.
\newblock \showarticletitle{Meal Delivery Routing Problem with a Hybrid Fleet
  of Riders and Autonomous Vehicles under Dynamic Environment}.
\newblock \bibinfo{journal}{\emph{IEEE Transactions on Automation Science and
  Engineering}} (\bibinfo{year}{2025}).
\newblock


\bibitem[NVIDIA(2025)]%
        {nvidia_cosmos_predict2}
\bibfield{author}{\bibinfo{person}{NVIDIA}.} \bibinfo{year}{2025}\natexlab{}.
\newblock \bibinfo{title}{COSMOS Predict2}.
\newblock
  \bibinfo{howpublished}{\url{https://github.com/nvidia-cosmos/cosmos-predict2}}.
\newblock
\newblock
\shownote{Accessed: 2025-08-26}.


\bibitem[OpenAI(2023)]%
        {sora2023}
\bibfield{author}{\bibinfo{person}{OpenAI}.} \bibinfo{year}{2023}\natexlab{}.
\newblock \bibinfo{title}{Sora: High-Fidelity Video Generation}.
\newblock \bibinfo{howpublished}{\url{https://openai.com/sora/}}.
\newblock
\newblock
\shownote{Accessed: 2025-05-30}.


\bibitem[Panowicz and Stecz(2024)]%
        {panowicz2024robust}
\bibfield{author}{\bibinfo{person}{Robert Panowicz} {and}
  \bibinfo{person}{Wojciech Stecz}.} \bibinfo{year}{2024}\natexlab{}.
\newblock \showarticletitle{Robust Optimization Models for Planning Drone Swarm
  Missions}.
\newblock \bibinfo{journal}{\emph{Drones}} \bibinfo{volume}{8},
  \bibinfo{number}{10} (\bibinfo{year}{2024}), \bibinfo{pages}{572}.
\newblock


\bibitem[Peebles and Xie(2023)]%
        {peebles2023scalable}
\bibfield{author}{\bibinfo{person}{William Peebles} {and}
  \bibinfo{person}{Saining Xie}.} \bibinfo{year}{2023}\natexlab{}.
\newblock \showarticletitle{Scalable diffusion models with transformers}. In
  \bibinfo{booktitle}{\emph{Proceedings of the IEEE/CVF international
  conference on computer vision}}. \bibinfo{pages}{4195--4205}.
\newblock


\bibitem[Rafael et~al\mbox{.}(2020)]%
        {rafael2020autonomous}
\bibfield{author}{\bibinfo{person}{Sandra Rafael}, \bibinfo{person}{Lu{\'\i}s~P
  Correia}, \bibinfo{person}{Diogo Lopes}, \bibinfo{person}{Jorge Bandeira},
  \bibinfo{person}{Margarida~C Coelho}, \bibinfo{person}{M{\'a}rio Andrade},
  \bibinfo{person}{Carlos Borrego}, {and} \bibinfo{person}{Ana~I Miranda}.}
  \bibinfo{year}{2020}\natexlab{}.
\newblock \showarticletitle{Autonomous vehicles opportunities for cities air
  quality}.
\newblock \bibinfo{journal}{\emph{Science of the Total Environment}}
  \bibinfo{volume}{712} (\bibinfo{year}{2020}), \bibinfo{pages}{136546}.
\newblock


\bibitem[Russell et~al\mbox{.}(2025)]%
        {russell2025gaia}
\bibfield{author}{\bibinfo{person}{Lloyd Russell}, \bibinfo{person}{Anthony
  Hu}, \bibinfo{person}{Lorenzo Bertoni}, \bibinfo{person}{George Fedoseev},
  \bibinfo{person}{Jamie Shotton}, \bibinfo{person}{Elahe Arani}, {and}
  \bibinfo{person}{Gianluca Corrado}.} \bibinfo{year}{2025}\natexlab{}.
\newblock \showarticletitle{Gaia-2: A controllable multi-view generative world
  model for autonomous driving}.
\newblock \bibinfo{journal}{\emph{arXiv preprint arXiv:2503.20523}}
  (\bibinfo{year}{2025}).
\newblock


\bibitem[Sakagami et~al\mbox{.}(2023)]%
        {sakagami2023robotic}
\bibfield{author}{\bibinfo{person}{Ryo Sakagami}, \bibinfo{person}{Florian~S
  Lay}, \bibinfo{person}{Andreas D{\"o}mel}, \bibinfo{person}{Martin~J
  Schuster}, \bibinfo{person}{Alin Albu-Sch{\"a}ffer}, {and}
  \bibinfo{person}{Freek Stulp}.} \bibinfo{year}{2023}\natexlab{}.
\newblock \showarticletitle{Robotic world models—conceptualization, review,
  and engineering best practices}.
\newblock \bibinfo{journal}{\emph{Frontiers in Robotics and AI}}
  \bibinfo{volume}{10} (\bibinfo{year}{2023}), \bibinfo{pages}{1253049}.
\newblock


\bibitem[Schedl et~al\mbox{.}(2021)]%
        {schedl2021autonomous}
\bibfield{author}{\bibinfo{person}{David~C Schedl}, \bibinfo{person}{Indrajit
  Kurmi}, {and} \bibinfo{person}{Oliver Bimber}.}
  \bibinfo{year}{2021}\natexlab{}.
\newblock \showarticletitle{An autonomous drone for search and rescue in
  forests using airborne optical sectioning}.
\newblock \bibinfo{journal}{\emph{Science Robotics}} \bibinfo{volume}{6},
  \bibinfo{number}{55} (\bibinfo{year}{2021}), \bibinfo{pages}{eabg1188}.
\newblock


\bibitem[Skorokhodov et~al\mbox{.}(2024)]%
        {skorokhodov2024hierarchical}
\bibfield{author}{\bibinfo{person}{Ivan Skorokhodov}, \bibinfo{person}{Willi
  Menapace}, \bibinfo{person}{Aliaksandr Siarohin}, {and}
  \bibinfo{person}{Sergey Tulyakov}.} \bibinfo{year}{2024}\natexlab{}.
\newblock \showarticletitle{Hierarchical patch diffusion models for
  high-resolution video generation}. In \bibinfo{booktitle}{\emph{Proceedings
  of the IEEE/CVF Conference on Computer Vision and Pattern Recognition}}.
  \bibinfo{pages}{7569--7579}.
\newblock


\bibitem[Teed and Deng(2020)]%
        {teed2020raft}
\bibfield{author}{\bibinfo{person}{Zachary Teed} {and} \bibinfo{person}{Jia
  Deng}.} \bibinfo{year}{2020}\natexlab{}.
\newblock \showarticletitle{Raft: Recurrent all-pairs field transforms for
  optical flow}. In \bibinfo{booktitle}{\emph{European conference on computer
  vision}}. Springer, \bibinfo{pages}{402--419}.
\newblock


\bibitem[Tot et~al\mbox{.}(2025)]%
        {tot2025adapting}
\bibfield{author}{\bibinfo{person}{Marko Tot}, \bibinfo{person}{Shu Ishida},
  \bibinfo{person}{Abdelhak Lemkhenter}, \bibinfo{person}{David Bignell},
  \bibinfo{person}{Pallavi Choudhury}, \bibinfo{person}{Chris Lovett},
  \bibinfo{person}{Luis Fran{\c{c}}a}, \bibinfo{person}{Matheus Ribeiro~Furtado
  de Mendon{\c{c}}a}, \bibinfo{person}{Tarun Gupta}, \bibinfo{person}{Darren
  Gehring}, {et~al\mbox{.}}} \bibinfo{year}{2025}\natexlab{}.
\newblock \showarticletitle{Adapting a World Model for Trajectory Following in
  a 3D Game}.
\newblock \bibinfo{journal}{\emph{arXiv preprint arXiv:2504.12299}}
  (\bibinfo{year}{2025}).
\newblock


\bibitem[Unterthiner et~al\mbox{.}(2018)]%
        {unterthiner2018towards}
\bibfield{author}{\bibinfo{person}{Thomas Unterthiner}, \bibinfo{person}{Sjoerd
  Van~Steenkiste}, \bibinfo{person}{Karol Kurach}, \bibinfo{person}{Raphael
  Marinier}, \bibinfo{person}{Marcin Michalski}, {and} \bibinfo{person}{Sylvain
  Gelly}.} \bibinfo{year}{2018}\natexlab{}.
\newblock \showarticletitle{Towards accurate generative models of video: A new
  metric \& challenges}.
\newblock \bibinfo{journal}{\emph{arXiv preprint arXiv:1812.01717}}
  (\bibinfo{year}{2018}).
\newblock


\bibitem[Wan et~al\mbox{.}(2025)]%
        {wan2025wan}
\bibfield{author}{\bibinfo{person}{Team Wan}, \bibinfo{person}{Ang Wang},
  \bibinfo{person}{Baole Ai}, \bibinfo{person}{Bin Wen},
  \bibinfo{person}{Chaojie Mao}, \bibinfo{person}{Chen-Wei Xie},
  \bibinfo{person}{Di Chen}, \bibinfo{person}{Feiwu Yu},
  \bibinfo{person}{Haiming Zhao}, \bibinfo{person}{Jianxiao Yang},
  {et~al\mbox{.}}} \bibinfo{year}{2025}\natexlab{}.
\newblock \showarticletitle{Wan: Open and advanced large-scale video generative
  models}.
\newblock \bibinfo{journal}{\emph{arXiv preprint arXiv:2503.20314}}
  (\bibinfo{year}{2025}).
\newblock


\bibitem[Wang et~al\mbox{.}(2025b)]%
        {wang2025ultra}
\bibfield{author}{\bibinfo{person}{Haoyang Wang}, \bibinfo{person}{Jingao Xu},
  \bibinfo{person}{Xinyu Luo}, \bibinfo{person}{Xuecheng Chen},
  \bibinfo{person}{Ting Zhang}, \bibinfo{person}{Ruiyang Duan},
  \bibinfo{person}{Yunhao Liu}, {and} \bibinfo{person}{Xinlei Chen}.}
  \bibinfo{year}{2025}\natexlab{b}.
\newblock \showarticletitle{Ultra-high-frequency harmony: mmwave radar and
  event camera orchestrate accurate drone landing}. In
  \bibinfo{booktitle}{\emph{Proceedings of the 23rd ACM Conference on Embedded
  Networked Sensor Systems}}. \bibinfo{pages}{15--29}.
\newblock


\bibitem[Wang et~al\mbox{.}(2024a)]%
        {wang2024transformloc}
\bibfield{author}{\bibinfo{person}{Haoyang Wang}, \bibinfo{person}{Jingao Xu},
  \bibinfo{person}{Chenyu Zhao}, \bibinfo{person}{Zihong Lu},
  \bibinfo{person}{Yuhan Cheng}, \bibinfo{person}{Xuecheng Chen},
  \bibinfo{person}{Xiao-Ping Zhang}, \bibinfo{person}{Yunhao Liu}, {and}
  \bibinfo{person}{Xinlei Chen}.} \bibinfo{year}{2024}\natexlab{a}.
\newblock \showarticletitle{Transformloc: Transforming mavs into mobile
  localization infrastructures in heterogeneous swarms}. In
  \bibinfo{booktitle}{\emph{IEEE INFOCOM 2024-IEEE Conference on Computer
  Communications}}. IEEE, \bibinfo{pages}{1101--1110}.
\newblock


\bibitem[Wang et~al\mbox{.}(2025a)]%
        {wang2025vggt}
\bibfield{author}{\bibinfo{person}{Jianyuan Wang}, \bibinfo{person}{Minghao
  Chen}, \bibinfo{person}{Nikita Karaev}, \bibinfo{person}{Andrea Vedaldi},
  \bibinfo{person}{Christian Rupprecht}, {and} \bibinfo{person}{David
  Novotny}.} \bibinfo{year}{2025}\natexlab{a}.
\newblock \showarticletitle{Vggt: Visual geometry grounded transformer}. In
  \bibinfo{booktitle}{\emph{Proceedings of the Computer Vision and Pattern
  Recognition Conference}}. \bibinfo{pages}{5294--5306}.
\newblock


\bibitem[Wang et~al\mbox{.}(2023)]%
        {wang2023videofactory}
\bibfield{author}{\bibinfo{person}{Wenjing Wang}, \bibinfo{person}{Huan Yang},
  \bibinfo{person}{Zixi Tuo}, \bibinfo{person}{Huiguo He},
  \bibinfo{person}{Junchen Zhu}, \bibinfo{person}{Jianlong Fu}, {and}
  \bibinfo{person}{Jiaying Liu}.} \bibinfo{year}{2023}\natexlab{}.
\newblock \showarticletitle{Videofactory: Swap attention in spatiotemporal
  diffusions for text-to-video generation}.
\newblock  (\bibinfo{year}{2023}).
\newblock


\bibitem[Wang et~al\mbox{.}(2024b)]%
        {wang2024drivedreamer}
\bibfield{author}{\bibinfo{person}{Xiaofeng Wang}, \bibinfo{person}{Zheng Zhu},
  \bibinfo{person}{Guan Huang}, \bibinfo{person}{Xinze Chen},
  \bibinfo{person}{Jiagang Zhu}, {and} \bibinfo{person}{Jiwen Lu}.}
  \bibinfo{year}{2024}\natexlab{b}.
\newblock \showarticletitle{DriveDreamer: Towards Real-World-Drive World Models
  for Autonomous Driving}. In \bibinfo{booktitle}{\emph{European Conference on
  Computer Vision}}. Springer, \bibinfo{pages}{55--72}.
\newblock


\bibitem[Wang et~al\mbox{.}(2024c)]%
        {wang2024worlddreamer}
\bibfield{author}{\bibinfo{person}{Xiaofeng Wang}, \bibinfo{person}{Zheng Zhu},
  \bibinfo{person}{Guan Huang}, \bibinfo{person}{Boyuan Wang},
  \bibinfo{person}{Xinze Chen}, {and} \bibinfo{person}{Jiwen Lu}.}
  \bibinfo{year}{2024}\natexlab{c}.
\newblock \showarticletitle{Worlddreamer: Towards general world models for
  video generation via predicting masked tokens}.
\newblock \bibinfo{journal}{\emph{arXiv preprint arXiv:2401.09985}}
  (\bibinfo{year}{2024}).
\newblock


\bibitem[Wu et~al\mbox{.}(2023)]%
        {wu2023daydreamer}
\bibfield{author}{\bibinfo{person}{Philipp Wu}, \bibinfo{person}{Alejandro
  Escontrela}, \bibinfo{person}{Danijar Hafner}, \bibinfo{person}{Pieter
  Abbeel}, {and} \bibinfo{person}{Ken Goldberg}.}
  \bibinfo{year}{2023}\natexlab{}.
\newblock \showarticletitle{Daydreamer: World models for physical robot
  learning}. In \bibinfo{booktitle}{\emph{Conference on robot learning}}. PMLR,
  \bibinfo{pages}{2226--2240}.
\newblock


\bibitem[Xu et~al\mbox{.}(2024)]%
        {xu2024scalable}
\bibfield{author}{\bibinfo{person}{Yanggang Xu}, \bibinfo{person}{Jirong Zha},
  \bibinfo{person}{Jiyuan Ren}, \bibinfo{person}{Xintao Jiang},
  \bibinfo{person}{Hongfei Zhang}, {and} \bibinfo{person}{Xinlei Chen}.}
  \bibinfo{year}{2024}\natexlab{}.
\newblock \showarticletitle{Scalable multi-agent reinforcement learning for
  effective uav scheduling in multi-hop emergency networks}. In
  \bibinfo{booktitle}{\emph{Proceedings of the 30th Annual International
  Conference on Mobile Computing and Networking}}. \bibinfo{pages}{2028--2033}.
\newblock


\bibitem[Yang et~al\mbox{.}(2024)]%
        {yang2024cogvideox}
\bibfield{author}{\bibinfo{person}{Zhuoyi Yang}, \bibinfo{person}{Jiayan Teng},
  \bibinfo{person}{Wendi Zheng}, \bibinfo{person}{Ming Ding},
  \bibinfo{person}{Shiyu Huang}, \bibinfo{person}{Jiazheng Xu},
  \bibinfo{person}{Yuanming Yang}, \bibinfo{person}{Wenyi Hong},
  \bibinfo{person}{Xiaohan Zhang}, \bibinfo{person}{Guanyu Feng},
  {et~al\mbox{.}}} \bibinfo{year}{2024}\natexlab{}.
\newblock \showarticletitle{Cogvideox: Text-to-video diffusion models with an
  expert transformer}.
\newblock \bibinfo{journal}{\emph{arXiv preprint arXiv:2408.06072}}
  (\bibinfo{year}{2024}).
\newblock


\bibitem[Ye et~al\mbox{.}(2022)]%
        {ye2022unsupervised}
\bibfield{author}{\bibinfo{person}{Junjie Ye}, \bibinfo{person}{Changhong Fu},
  \bibinfo{person}{Guangze Zheng}, \bibinfo{person}{Danda~Pani Paudel}, {and}
  \bibinfo{person}{Guang Chen}.} \bibinfo{year}{2022}\natexlab{}.
\newblock \showarticletitle{Unsupervised domain adaptation for nighttime aerial
  tracking}. In \bibinfo{booktitle}{\emph{Proceedings of the IEEE/CVF
  conference on computer vision and pattern recognition}}.
  \bibinfo{pages}{8896--8905}.
\newblock


\bibitem[Yu et~al\mbox{.}(2022)]%
        {yu2022scaling}
\bibfield{author}{\bibinfo{person}{Jiahui Yu}, \bibinfo{person}{Yuanzhong Xu},
  \bibinfo{person}{Jing~Yu Koh}, \bibinfo{person}{Thang Luong},
  \bibinfo{person}{Gunjan Baid}, \bibinfo{person}{Zirui Wang},
  \bibinfo{person}{Vijay Vasudevan}, \bibinfo{person}{Alexander Ku},
  \bibinfo{person}{Yinfei Yang}, \bibinfo{person}{Burcu~Karagol Ayan},
  {et~al\mbox{.}}} \bibinfo{year}{2022}\natexlab{}.
\newblock \showarticletitle{Scaling autoregressive models for content-rich
  text-to-image generation}.
\newblock \bibinfo{journal}{\emph{arXiv preprint arXiv:2206.10789}}
  \bibinfo{volume}{2}, \bibinfo{number}{3} (\bibinfo{year}{2022}),
  \bibinfo{pages}{5}.
\newblock


\bibitem[Zha et~al\mbox{.}(2025)]%
        {zha2025enable}
\bibfield{author}{\bibinfo{person}{Jirong Zha}, \bibinfo{person}{Yuxuan Fan},
  \bibinfo{person}{Xiao Yang}, \bibinfo{person}{Chen Gao}, {and}
  \bibinfo{person}{Xinlei Chen}.} \bibinfo{year}{2025}\natexlab{}.
\newblock \showarticletitle{How to enable llm with 3d capacity? a survey of
  spatial reasoning in llm}.
\newblock \bibinfo{journal}{\emph{arXiv preprint arXiv:2504.05786}}
  (\bibinfo{year}{2025}).
\newblock


\bibitem[Zhang et~al\mbox{.}(2023)]%
        {10004511}
\bibfield{author}{\bibinfo{person}{Chunhui Zhang}, \bibinfo{person}{Guanjie
  Huang}, \bibinfo{person}{Li Liu}, \bibinfo{person}{Shan Huang},
  \bibinfo{person}{Yinan Yang}, \bibinfo{person}{Xiang Wan},
  \bibinfo{person}{Shiming Ge}, {and} \bibinfo{person}{Dacheng Tao}.}
  \bibinfo{year}{2023}\natexlab{}.
\newblock \showarticletitle{WebUAV-3M: A Benchmark for Unveiling the Power of
  Million-Scale Deep UAV Tracking}.
\newblock \bibinfo{journal}{\emph{IEEE Transactions on Pattern Analysis and
  Machine Intelligence}} \bibinfo{volume}{45}, \bibinfo{number}{7}
  (\bibinfo{year}{2023}), \bibinfo{pages}{9186--9205}.
\newblock
\href{https://doi.org/10.1109/TPAMI.2022.3232854}{doi:\nolinkurl{10.1109/TPAMI.2022.3232854}}


\bibitem[Zhang et~al\mbox{.}(2025)]%
        {zhang2025citynavagent}
\bibfield{author}{\bibinfo{person}{Weichen Zhang}, \bibinfo{person}{Chen Gao},
  \bibinfo{person}{Shiquan Yu}, \bibinfo{person}{Ruiying Peng},
  \bibinfo{person}{Baining Zhao}, \bibinfo{person}{Qian Zhang},
  \bibinfo{person}{Jinqiang Cui}, \bibinfo{person}{Xinlei Chen}, {and}
  \bibinfo{person}{Yong Li}.} \bibinfo{year}{2025}\natexlab{}.
\newblock \showarticletitle{CityNavAgent: Aerial Vision-and-Language Navigation
  with Hierarchical Semantic Planning and Global Memory}.
\newblock \bibinfo{journal}{\emph{arXiv preprint arXiv:2505.05622}}
  (\bibinfo{year}{2025}).
\newblock


\bibitem[Zhao et~al\mbox{.}(2025a)]%
        {zhao2025urbanvideobench}
\bibfield{author}{\bibinfo{person}{Baining Zhao}, \bibinfo{person}{Jianjie
  Fang}, \bibinfo{person}{Zichao Dai}, \bibinfo{person}{Ziyou Wang},
  \bibinfo{person}{Jirong Zha}, \bibinfo{person}{Weichen Zhang},
  \bibinfo{person}{Chen Gao}, \bibinfo{person}{Yue Wang},
  \bibinfo{person}{Jinqiang Cui}, \bibinfo{person}{Xinlei Chen}, {and}
  \bibinfo{person}{Yong Li}.} \bibinfo{year}{2025}\natexlab{a}.
\newblock \showarticletitle{{U}rban{V}ideo-Bench: Benchmarking Vision-Language
  Models on Embodied Intelligence with Video Data in Urban Spaces}. In
  \bibinfo{booktitle}{\emph{Proceedings of the 63rd Annual Meeting of the
  Association for Computational Linguistics (Volume 1: Long Papers)}},
  \bibfield{editor}{\bibinfo{person}{Wanxiang Che}, \bibinfo{person}{Joyce
  Nabende}, \bibinfo{person}{Ekaterina Shutova}, {and}
  \bibinfo{person}{Mohammad~Taher Pilehvar}} (Eds.).
  \bibinfo{publisher}{Association for Computational Linguistics},
  \bibinfo{address}{Vienna, Austria}, \bibinfo{pages}{32400--32423}.
\newblock
\showISBNx{979-8-89176-251-0}
\href{https://doi.org/10.18653/v1/2025.acl-long.1558}{doi:\nolinkurl{10.18653/v1/2025.acl-long.1558}}


\bibitem[Zhao et~al\mbox{.}(2025b)]%
        {zhao2025embodiedr}
\bibfield{author}{\bibinfo{person}{Baining Zhao}, \bibinfo{person}{Ziyou Wang},
  \bibinfo{person}{Jianjie Fang}, \bibinfo{person}{Chen Gao},
  \bibinfo{person}{Fanhang Man}, \bibinfo{person}{Jinqiang Cui},
  \bibinfo{person}{Xin Wang}, \bibinfo{person}{Xinlei Chen},
  \bibinfo{person}{Yong Li}, {and} \bibinfo{person}{Wenwu Zhu}.}
  \bibinfo{year}{2025}\natexlab{b}.
\newblock \bibinfo{title}{Embodied-R: Collaborative Framework for Activating
  Embodied Spatial Reasoning in Foundation Models via Reinforcement Learning}.
\newblock
\showeprint[arxiv]{2504.12680}~[cs.AI]
\urldef\tempurl%
\url{https://arxiv.org/abs/2504.12680}
\showURL{%
\tempurl}


\bibitem[Zhao et~al\mbox{.}(2024)]%
        {zhao2024identifying}
\bibfield{author}{\bibinfo{person}{Min Zhao}, \bibinfo{person}{Hongzhou Zhu},
  \bibinfo{person}{Chendong Xiang}, \bibinfo{person}{Kaiwen Zheng},
  \bibinfo{person}{Chongxuan Li}, {and} \bibinfo{person}{Jun Zhu}.}
  \bibinfo{year}{2024}\natexlab{}.
\newblock \showarticletitle{Identifying and solving conditional image leakage
  in image-to-video diffusion model}.
\newblock \bibinfo{journal}{\emph{arXiv preprint arXiv:2406.15735}}
  (\bibinfo{year}{2024}).
\newblock


\bibitem[Zhou et~al\mbox{.}(2025)]%
        {zhou2025catua}
\bibfield{author}{\bibinfo{person}{Nan Zhou}, \bibinfo{person}{Yuxuan Liu},
  \bibinfo{person}{Haoyang Wang}, \bibinfo{person}{Fanhang Man},
  \bibinfo{person}{Jingao Xu}, \bibinfo{person}{Fan Dang},
  \bibinfo{person}{Chaopeng Hong}, \bibinfo{person}{Yunhao Liu},
  \bibinfo{person}{Xiao-Ping Zhang}, \bibinfo{person}{Yali Song},
  {et~al\mbox{.}}} \bibinfo{year}{2025}\natexlab{}.
\newblock \showarticletitle{CatUA: Catalyzing Urban Air Quality Intelligence
  through Mobile Crowd-sensing}.
\newblock \bibinfo{journal}{\emph{IEEE Transactions on Mobile Computing}}
  (\bibinfo{year}{2025}).
\newblock


\bibitem[Zhou et~al\mbox{.}(2024)]%
        {zhou2024robodreamer}
\bibfield{author}{\bibinfo{person}{Siyuan Zhou}, \bibinfo{person}{Yilun Du},
  \bibinfo{person}{Jiaben Chen}, \bibinfo{person}{Yandong Li},
  \bibinfo{person}{Dit-Yan Yeung}, {and} \bibinfo{person}{Chuang Gan}.}
  \bibinfo{year}{2024}\natexlab{}.
\newblock \showarticletitle{Robodreamer: Learning compositional world models
  for robot imagination}.
\newblock \bibinfo{journal}{\emph{arXiv preprint arXiv:2404.12377}}
  (\bibinfo{year}{2024}).
\newblock


\end{thebibliography}
	
	
\end{document}